\documentclass[12pt,twoside,a4paper]{article}

\usepackage{titlesec} % For section formatting and counter handling
\usepackage{ifxetex}
\usepackage{textpos}
\usepackage{kpfonts}
\usepackage[a4paper,hmargin=2.0cm,vmargin=2.0cm,includeheadfoot]{geometry}
\usepackage{setspace}

\usepackage{ifxetex}
\usepackage{stackengine}
\usepackage{tabularx,longtable,multirow,subfigure,caption}%hangcaption
\usepackage{fancyhdr}
\usepackage{color}
\usepackage[tight,ugly]{units}
\usepackage{url}
\usepackage[english]{babel}
\usepackage{amsmath}
\usepackage{graphicx}
\usepackage[colorinlistoftodos]{todonotes}
\usepackage{dsfont}
\usepackage{epstopdf} % automatically replace .eps with .pdf in graphics
\usepackage{array}
\usepackage{latexsym}
\usepackage{etoolbox}
\usepackage{hyperref}
\usepackage{enumerate} % for numbering with [a)] format 
\usepackage{algorithm}
\usepackage{algorithmic}

\usepackage{abstract}

\ifxetex
\usepackage{fontspec}
\else

\usepackage{ntheorem}
\theoremstyle{break}

\ifxetex
\else
\fi

\def\@makechapterhead#1{%
  \vspace*{10\p@}%
  {\parindent \z@ \raggedright %\sffamily
    \interlinepenalty\@M
    \Huge \bfseries 
    \thechapter \space\space #1\par\nobreak
    \vskip 30\p@
  }}

\def\@makeschapterhead#1{%
  \vspace*{10\p@}%
  {\parindent \z@ \raggedright
    \sffamily
    \interlinepenalty\@M
    \Huge \bfseries  
    #1\par\nobreak
    \vskip 30\p@
  }}

\def\Beginboxit
   {\par
    \vbox\bgroup
	   \hrule
	   \hbox\bgroup
		  \vrule \kern1.2pt %
		  \vbox\bgroup\kern1.2pt
   }

\def\Endboxit{%
			      \kern1.2pt
		       \egroup
		  \kern1.2pt\vrule
		\egroup
	   \hrule
	 \egroup
   }

\newenvironment{boxit*}{\Beginboxit\hbox to\hsize{}}{\Endboxit}

\allowdisplaybreaks

\makeatletter
\newcounter{elimination@steps}
\newcolumntype{R}[1]{>{\raggedleft\arraybackslash$}p{#1}<{$}}
\def\elimination@num@rights{}
\def\elimination@num@variables{}
\def\elimination@col@width{}

\newcommand{\eliminationstep}[2]
{
    \ifnum\value{elimination@steps}>0\leadsto\quad\fi
    \left[
        \ifnum\elimination@num@rights>0
            \begin{array}
            {@{}*{\elimination@num@variables}{R{\elimination@col@width}}
            |@{}*{\elimination@num@rights}{R{\elimination@col@width}}}
        \else
            \begin{array}
            {@{}*{\elimination@num@variables}{R{\elimination@col@width}}}
        \fi
            #1
        \end{array}
    \right]
    & 
    \begin{array}{l}
        #2
    \end{array}
    &%                                    moved second & here
    \addtocounter{elimination@steps}{1}
}
\makeatother

\makeatletter  
\def\colvec#1{\expandafter\colvec@i#1,,,,,,,,,\@nil}
\def\colvec@i#1,#2,#3,#4,#5,#6,#7,#8,#9\@nil{% 
  \ifx$#2$ \begin{bmatrix}#1\end{bmatrix} \else
    \ifx$#3$ \begin{bmatrix}#1\\#2\end{bmatrix} \else
      \ifx$#4$ \begin{bmatrix}#1\\#2\\#3\end{bmatrix}\else
        \ifx$#5$ \begin{bmatrix}#1\\#2\\#3\\#4\end{bmatrix}\else
          \ifx$#6$ \begin{bmatrix}#1\\#2\\#3\\#4\\#5\end{bmatrix}\else
            \ifx$#7$ \begin{bmatrix}#1\\#2\\#3\\#4\\#5\\#6\end{bmatrix}\else
              \ifx$#8$ \begin{bmatrix}#1\\#2\\#3\\#4\\#5\\#6\\#7\end{bmatrix}\else
                 \PackageError{Column Vector}{The vector you tried to write is too big, use bmatrix instead}{Try using the bmatrix environment}
              \fi
            \fi
          \fi
        \fi
      \fi
    \fi
  \fi 
}  
\makeatother

\robustify{\colvec}

\begin{document}

\begin{titlepage}
                % \newgeometry{top=25mm,bottom=25mm,left=38mm,right=32mm}
                \setlength{\parindent}{0pt}
                \setlength{\parskip}{0pt}
                % \fontfamily{phv}\selectfont

                {
                                \Large
                                \raggedright
                                Imperial College London\\[17pt]
                                Department of Electrical and Electronic Engineering\\[17pt]
                                Final Year Project Report 2019\\[17pt]
 
                }

                \rule{\columnwidth}{3pt}
                \vfill
                \centering
                    \includegraphics[width=0.5\textwidth]{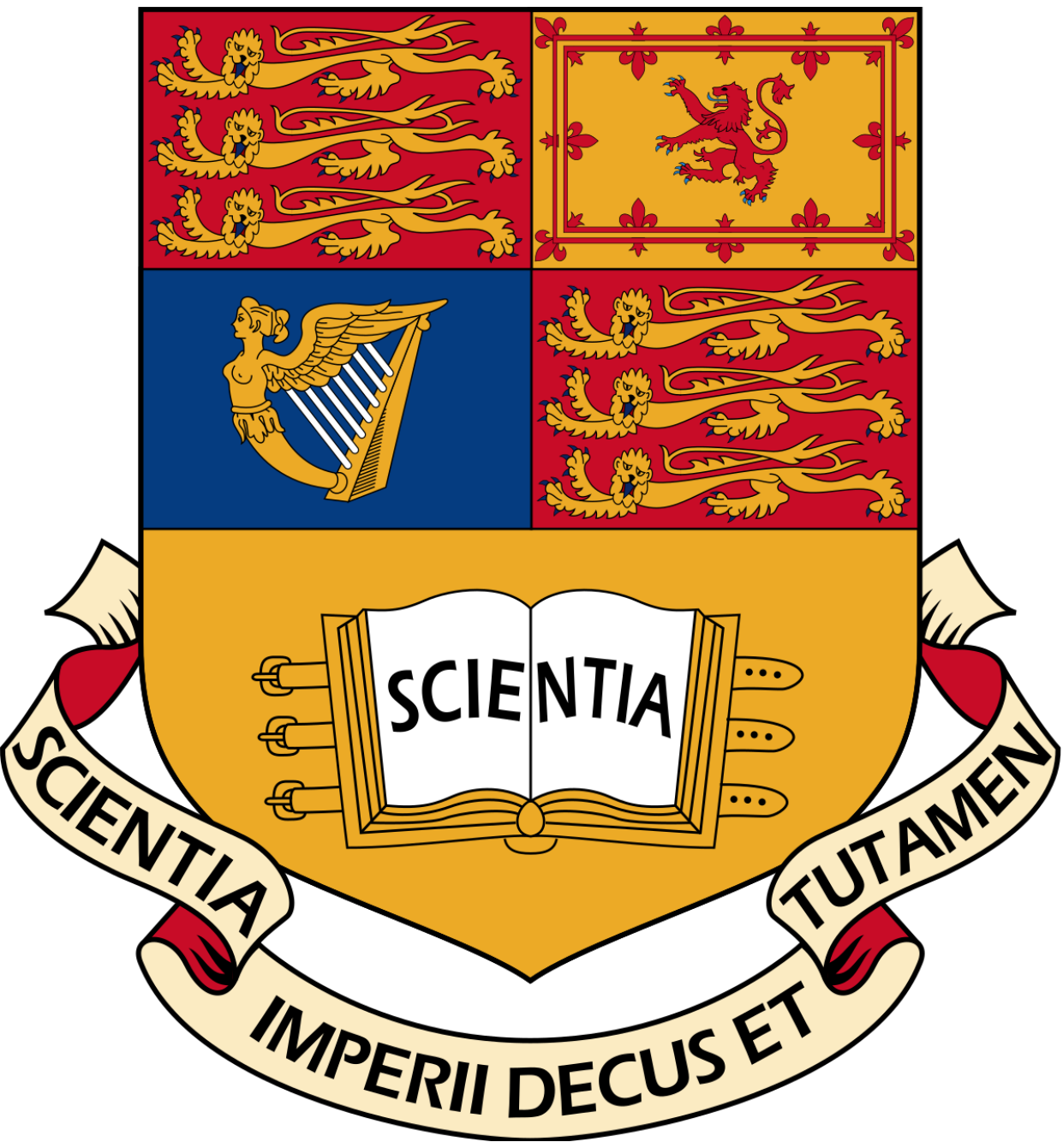}
                \vfill
                \setlength{\tabcolsep}{0pt}

                \begin{tabular}{p{40mm}p{\dimexpr\columnwidth-40mm}}
                                Project Title: & \textbf{Blind Source Separation Based on Sparsity} \\[12pt]
                                Student: & \textbf{Zhongxuan Li} \\[12pt]
                                CID: & \textbf{01050018} \\[12pt]
                                Course: & \textbf{EE4} \\[12pt]
                                Project Supervisor: & \textbf{Dr Wei Dai} \\[12pt]
                                Second Marker: & \textbf{Dr Deniz Gunduz} \\
                \end{tabular}
\end{titlepage}
\begin{abstract}
\noindent
Blind source separation (BSS) is a prevalent technique in array processing and data analysis that aims to recover unknown sources from observed mixtures, where the mixing matrix is unknown. The classical ICA methods require prior statistical knowledge that the sources have to be mutually independent. In order to overcome the limitations, sparsity based methods are introduced, which decompose the source signal sparsely in a prescribed dictionary. Morphological component analysis (MCA) theory is proposed based on theory of sparse representation. It assumes that the signal is a linear combination of several components having different geometry, and each embodiment of the component can be sparsely represented in a dictionary, and not sparsely represented in others. In recent years, this theory has been applied to solve the blind source separation problem and obtained good results.\\

The objectives of this report are to review some of the key approaches derived from the classical ICA methods that have been developed to address the BSS problem, and to further discuss sparsity based methods in blind source separation. It first describes the theory behind sparse representation and sparse decomposition algorithms, after which this report gives a decomposition algorithm based on block coordinate relaxation morphological component analysis whose variants have been applied to the multichannel morphological component analysis (MMCA) and generalised morphological component analysis (GMCA). A local dictionary learning (K-SVD) BSS algorithm is followed. Finally we improve the K-SVD BSS algorithm by further learning a block sparsifying dictionary (SAC+BK-SVD), which clusters the dictionary atoms according to their similarity and those atoms are updated by blocks. \\

In the implementation part, we are expected to perform image segmentation experiment and blind image source separation experiment using the techniques we have introduced. Another experiment involves comparing the proposed block-sparse dictionary learning algorithm with the K-SVD algorithm. Simulation results show the proposed methods yields better blind image separation quality. 
\end{abstract}
\newpage
\tableofcontents
\newpage

\section{Introduction} \label{intro}
\subsection{BSS preview}
Imagine that two people are speaking simultaneously in a room. There are two microphones which are in different locations and record the stereo signals generated by two people. Assuming that the each of the recorded time signals $x_1(t)$ and $x_2(t)$ is a linear combination of the speeches $s_1(t)$ and $s_2(t)$. We could express this as a set of linear equations:

\begin{align}
    x_1(t) = a_{11}s_1(t) + a_{12}s_2(t)\\
    x_2(t) = a_{21}s_1(t) + a_{22}s_2(t)
\end{align}

where parameter $a_{ij}$ depends on the distance of the microphones from the speakers. Time delays or other extra factors are ignored temporarily from our simplified mixing model. It would be useful in real-world applications if we can estimate the original speech signals $s_1(t)$ and $s_2(t)$ using only the recorded signals $x_1(t)$ and $x_2(t)$. This is addressed as the well-known cocktail party problem and also the chief motivation behind blind source separation. The similar situation is also common in telecomms, medical signal and image processing. If we knew the parameters $a_{ij}$ , we find $s_1(t)$ and $s_2(t)$  by solving linear equations or matrix inversion. The problem is, however we have no information about $a_{ij}$. the problem is considerable difficult.\\

The lack of prior knowledge of the mixing process can be compensated by a statistically strong but often physical plausible assumption of independence between the source signals. Independent Component Analysis (ICA) was first proposed to solve the cocktail party problem. The goal of ICA is to determine the original sources given mixtures of those sources, assuming that the sources are statistically independent and non Gaussian. Derivatives of the classical ICA methods includes JADE \cite{JADE720250}, FastICA \cite{fastICA777510}. Generally speaking, ICA algorithms are about devising adequate contrast functions which are related to approximation of independence\cite{HYVARINEN2000411}. However ICA is limited to the determined BSS problem when we have equal number of mixtures and the number of sources. This is because we need to find the inverse of the mixing matrix while optimising the contrast function in ICA. But only square matrix has such an inverse.

\subsection{BSS by sparsity}
Although ICA is proved to be effective in many BSS applications. The statistical independence assumption in the time domain cannot be applied to all scenarios. Sparsity-based approaches have drawn much attention in recent years. The term sparse refers to signals with small number of nonzeros with respect to some representation bases \cite{ZibulevskyMichael2001BSSb}. More specifically, sources have mutually disjoint support sets in a dictionary. This is exploited for instance in sparse component analysis (SCA) \cite{SCA2005}. In SCA we make assumption that the sources to be unmixed can be sparsely represented in a predefined common basis or dictionary (for instance, a wavelet frame). A two-step approach \cite{BOFILL20012353} was proposed to solve the BSS problem using sparsity, in which the mixing system is first estimated using clustering methods, then the sources are estimated thanks to pursuit methods (e.g. basis pursuit, matching pursuit).\\

In many cases, basis pursuit or matching pursuit synthesis algorithms are computationally quite expensive. Furthermore, the traditional SCA requires highly sparse signals. Unfortunately, this is not the case for high dimensional signals and especially in image processing. We present in this report an alternative to these approaches, the morphological component analysis (MCA) \cite{BobinJ_2007SaMD, BobinJ_2006Mdas} is a method which sources can be sparsely represented using several different dictionaries. For example, images normally contains contour and texture, the former is well sparsified using curvelets transform whereas the latter may be well represented using local cosine transform (DCT). 
%Extensions of MCA have been proposed to solve the case when each distinct source can be sparse representated in a specific dictionary. 
Multichannel morphological component analysis (MMCA) \cite{Starck2005MorphologicalCA} and generalised morphological component analysis (GMCA) are extentions of MCA to the multichannel case. In MMCA setting, we assume that the sources have strictly different morphologies (i.e., each source is assumed to be sparsely representated in one particular orthonormal basis). In GMCA, each source is modeled as the linear combination of a number of morphologial components where each component is sparse in a specific orthonormal basis. 

\subsection{Structure of report}
In next chapter, we first provide sufficient background knowledge for the blind source separation problem setting. Real world applications of BSS will be introduced before the performance measurements to evaluate different BSS algorithms are defined. Moreover, we will discuss the the well-established ICA algorithm as it is selected as our baseline method. We then turn our discussion to sparsity and morphological diversity. The idea of overcomplete dictionaries is followed. We will look in to multichannel morphological component Analysis and generalised morphological component analysis. In addition blind source separation based on adaptive dictionary learning is introduced. In Chapter 3, we run MATLAB simulations using the method discussed. Finally, we look into the future research directions in blind source separation.

\section{Background} \label{background}
\subsection{Instantaneous linear mixture model} \label{bssmodel}
Based on the cocktail party problem introduced in Chapter \ref{intro}, here we extent the idea of blind source separation to a formal mathematical definition. %Before that, we simply the BSS problem by applying the instantaneously linear mixture model.
The mixing process of the sources in BSS involves many models such as the instantaneous linear mixture model, the nonlinear mixture model and the convolved mixture model \cite{Hxu2014}. The instantaneous linear model omits the time delay of source propagation of reaching different observers. We assume that the instantaneous linear BSS model is adopted throughout this report.\\

\begin{figure}[H]
\centering
\includegraphics[width=0.45\textwidth]{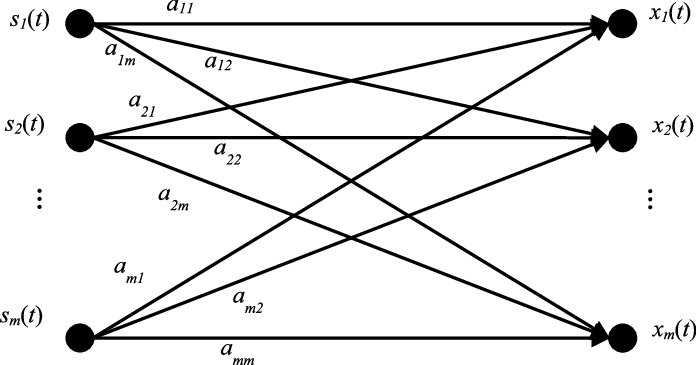}
\caption{Instantaneous linear mixture model (m=n)}
\label{Ins_LMM}
\end{figure}

The instantaneously linear mixture model states that given $m$ observations $\{x_1,\cdots, x_m\}$ where each $\{x_i\}_{i=1, \cdots,m}$ is a row vector of size $t$. Each observation is the linear mixture of $n$ sources $\{s_1^T,\cdots, s_n^T\}$ weighted by $a_{ij}$

\begin{equation}
    \forall i \in \{1, \cdots, m\}, \quad x_i = \sum_{j=1}^n a_{ij}s_j
    \label{sumofs_t}
\end{equation}
Figure (\ref{Ins_LMM}) illustrates the case when we have equal number of sources and observations. Since results are not affected by reciprocal rescaling of $a_{ij}$ and $s_j$. Without loss of generality, the $a_{ij}$ will hitherto assumed to be normalised to unit length. The mixing model can be conveniently rewritten in matrix form
\begin{equation}
    \mathbf{X} = \mathbf{AS} +\mathbf{N}   
    \label{mixing}
\end{equation}

where $\mathbf{X}$ is the observation matrix with dimension $m \times t$, $\mathbf{S}$ is the $n\times t$ source matrix and $\mathbf{A}$ is the $m \times n$ mixing matrix. An $m \times t$ matrix $\mathbf{N}$ accounts for additive noise or model imperfections. Under the blind separation problem setting, both $\mathbf{A}$ and $\mathbf{S}$ are unknown.  %$\mathbf{A}$ will be assumed to be full rank when ICA is in use. 
Source separation techniques aim at recovering the original signal $\mathbf{S} = [s_1^T,\cdots, s_n^T]$ from $m$ different mixtures by taking advantage of some information in the way the signals are mixed in observed data. In other words, source separation simply boils down to devising quantitative measures of diversity or contrast to differentiate between the sources.\\

Mathematically, We aim to find a demixing matrix $\mathbf{W}$ with dimension $n \times m$ which gives a linear combination of columns in the  observation $\mathbf{X}$, omitting the noise for now, that is
\begin{equation}
    \mathbf{Y} = \mathbf{WX}
    \label{demixing}
\end{equation}

$\mathbf{Y}$ is hence an estimation of source $\mathbf{S}$.
Combining Eq. (\ref{mixing}) and (\ref{demixing}) $y$ can be written as
\begin{equation}
    \mathbf{Y} = \mathbf{W^T} \mathbf{X} = \mathbf{W A S} = \mathbf{Z^T S}
    \label{estimationYandZ}
\end{equation}
Where $\mathbf{Z^T} = \mathbf{W A}$. The estimation matrix $\mathbf{Y}$ can also be represented as 
\begin{equation}
    \mathbf{Y} = \mathbf{PDS} 
    \label{permutEQ}
\end{equation}
where $\mathbf{D}$ is a diagonal matrix and $\mathbf{P}$ is a permutation matrix that reduces the scale permutation indeterminacy of the mixing model.

\subsection{Ambiguities of BSS process}
\label{ambiguitiesBSS}
Because lack of prior knowledge about the sources and mixing process. While estimating the mixing and source matrices, we introduce the diagonal matrix $\mathbf{D}$ and permutation matrix $\mathbf{P}$ in Equation (\ref{permutEQ}), which accounts for the two ambiguities, amplitude and order.\\

Variances (energies) of each source component is unknown. The reason is that, both $\mathbf{S}$ and $\mathbf{A}$ are not defined, any scalar multiplication in one of the sources $s_i$ would always be cancelled by dividing the corresponding column $a_i$ by the same quantity. However, the information of sources is stored in the signal waveform rather than amplitude. This ambiguity is hence insignificant in most applications \cite{HYVARINEN2000411}. As mentioned in the last section, the columns of $\mathbf{A}$ is normalised to unity just for convenience in calculation.\\

Apart from amplitude uncertainty, we cannot determine the order of the sources. In Equation (\ref{sumofs_t}), we can freely rearrange the sources up to any permutation without affecting the observation samples. Any of the source components can be regarded as the first one. Fortunately, we can use certain techniques (e.g. Hungarian Algorithm) to sort the recovered sources after separation has been done. Again, because the information of each component is contained in the shape of waveform of that component, not the order of components. So this ambiguity is also insignificant.

\subsection{Preprocessing of BSS techniques}
Before applying the BSS methods on the data, it is usually very usful to do some preprocessing. In this section, we introduce two preprocessing techniques that make the blind source separation problem simpler and better conditioned.
1. Centering:
Most BSS methods assumes the data to be zero centered. The most basic and necessary preprocessing is to center the observation data $x_i$ by substract the sample mean from it. 
2. Whitening:
Before applying BSS methods (and after centering), the observed vector $x$ is linearly transformed to $\tilde{x}$ so that each column is uncorrelated and have unity variance. More specifically, the covariance matrix of $\tilde{x}$ equals the identity martix, $\mathbb{E}\left\{\tilde{x}\tilde{x}^T\right\} = \mathbf{I}$. The most common used whitening transformation is eigenvalues decompostion (EVD) of the data covariance matrix.

\subsection{BSS performance measures}
\label{perform_metric}
1. Correlation coefficient measures the similarity between a recovered source $S^{'}$ and the original source $S$. Larger correlation coefficients indicate the original sources are better recovered.
\begin{equation}
    \rho = \frac{\text{cov}(S^{'},S)}{\sigma_x \sigma_y}
\end{equation}

2. Mixing Matrix Criterion assesses the separation quality due to demixing matrix $\mathbf{A}$, especially in noisy content.
\begin{equation}
    C_A = ||\mathbf{I_n} - P\tilde{\mathbf{A}}^{+}\mathbf{A} ||
\end{equation}

where $\mathbf{I}$ is the identity matrix, $\mathbf{P}$ is the permutation matrix, and $\tilde{\mathbf{A}}^{+}$ is the pseudo-inverse of the estimated mixing matrix. The mixing matrix criterion is strictly positive, unless the mixing matrix is correctly estimated up to scale and permutation \cite{VAbolghasemi2012}. Low values of $C_A$ then indicate better separation performance.
 
3. Human visual system (HVS): Most of out work in this report focus on blind image separation. Human visual system says that people are not as sensitive to high frequency detail as to low frequency ones. Therefore, standard metrics may not best describe the actual experiment outcomes. In order to better evaluate the results in image processing, we adopt HVS as the subjective metric.

\subsection{Applications of BSS}
The classical application of BSS on the cocktail party problem is trying to understand how the humans select the voice of a particular speaker from an ensemble of different voices corrupted by music and noise in the background. Other applications, besides the cocktail party problem mentioned in the introduction, have also attracted researchers' attention in the past decade. Examples are given as below. \\

An electroencephalogram (EEG) is a test used to find problems related to electrical activity of the brain. In EEG analysis, different artifacts such as eye-blinking deteriorate its quality. Identification of the various sources from the independent components is thus integral for clinical analysis. An innovative method combining the use of standard BSS techniques and Support Vector Machines (SVM) was applied to solve this problem  \cite{Duda2000PC954544}.\\

\begin{figure}[!htbp]
\centering
\includegraphics[width=0.45\textwidth]{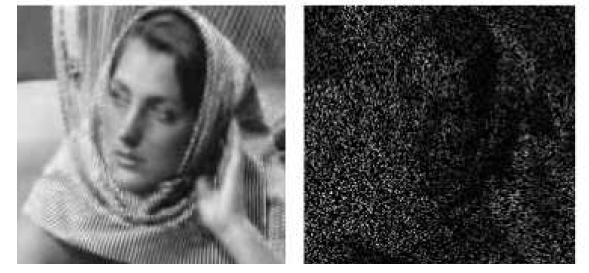}
\caption{Barbara image with $80\%$ missing pixels (right). The result of the MCA impainting is
given on the left.}
\label{imapint_1}
\end{figure}

Filling `holes' in images is an interested and important inverse problem with applications in repairing the old and deteriorated artwork. Based on Morphological Component Analysis, an inpainting algorithm has been proposed  which is capable of filling holes in either texture or cartoon content \cite{ELAD2005340}. In Figure (\ref{imapint_1}), the inpainting algorithm is applied on the famous Barbara images and achieve satisfactory result even when there are 80\% missing pixels. \\

In Code-Division Multiple Access (CDMA), blind separation techniques are used to suppress unintentional multiuser interference (jammer), separate the desired user signals from other users' signals \cite{Raju2006}. BSS also has wide applications in military based telecomm systems which recovers radar reflection from strong intentional interference.\\

BSS is also closely related with studying the underlying factors of the financial data and driving mechanisms behind financial time series. In \cite{OjaE2000Icaf} ICA is applied to financial time series data. The data is parallel, representing the simultaneous cash flow at several stores belonging to the same retail chain. The ICA finds the fundamental factors that are common to all stores that affect the cashflow data, although each store responds to these factors in a slightly different manner. Thus, the cashflow effect of the factors specific to any particular store could be revealed.

\subsection{Independent component analysis}
ICA algorithms are about devising appropriate independence approximations. This includes, maximisation of non-Gaussianity, minimisation of mutual information and maximum likelihood estimation \cite{HYVARINEN2000411}. We are not going to discuss all these approximations in depth as we want to focus more on the sparsity based blind source separation. Only the FastICA and JADE are introduced below.\\

FastICA calculates the negentropy as an approximate of independence measure. With a taste of the central limit theorem, intuition tells us that the distribution of a sum of independent random variables tends to toward a Gaussian distribution, under certain conditions. Known from Equation (\ref{estimationYandZ}) that $y = w^T x = w A s = z^T s$. It is clear that the closest estimation is when $z^Ts = s$ which also has the least Gaussianity. Hence finding independent $s$ is equivalent to minimisation of Gaussianity. The approximation of non-Gaussianity is based on a maximum negentropy principle.
\begin{equation}
    J(y) \propto [E\{G(y)\} - E\{G(v)\}]^2
\end{equation}
Where $v$ is a standardised Guassian variable. And $G$ are predefined functions with the possible choices stated in \cite{HYVARINEN2000411}. In general, FastICA is based on a fixed point iteration scheme for finding the maximum of non-Gaussianity of $\mathbf{w^T x}$ in Equation (\ref{demixing}). The basic form of FastICA algorithm is as follows.

% -------------------- algorithm block ------------------------
\begin{algorithm}[!htbp] 
\caption{ The basic FastICA algorithm for estimating one independent component}
\label{alg:Framwork} 
\begin{algorithmic}
\REQUIRE ~~\\%Input
The observed matrix $\mathbf{x}$
\ENSURE ~~\\ %Output
Estimation of $\hat{\mathbf{A}}$
\STATE 1. Choose an initial (e.g. random) wright vector $\mathbf{w}$
\STATE 2. Let $\mathbf{w}^+ = E\{\mathbf{x}g(w^T x) - E(g'(w^Tx)\}\mathbf{w})$
\STATE 3. Calculate $\mathbf{w} = \mathbf{w}^{+}/\lVert w^{+}\rVert$
\STATE 4. If not converged (old and new values of $\mathbf{w}$ point in the same direction up to multiplicative signs), return to step 2
\end{algorithmic}
\end{algorithm}
% -------------------- algorithm block ------------------------

JADE is based on similar ideas of the FastICA agorithm apart from it calculates the fourth-order statistics (Kurtosis). JADE finds out the direction where the kurtosis of observed signal grows most strongly (super-Gaussian signals) or decreases most strongly (sub-Gaussian signals).\\

\label{ica_defect}
ICA suffers from several limitations which make it unsuitable in some specific applications. Firstly, ICA require the mixing matrix $\mathbf{A}$ to be full rank and square. As we mentioned in the introdcution, generally ICA cannot be applied to the underdetermined mixing scenario. Secondly, ICA also assumes that amongst the components in $\mathbf{S}$, there exists at most one component that is Gaussian. This means ICA is not robust under the additive Gaussian nosie setting. While even implicit, the ICA algorithm requires information on the source distribution when doing separating computation such as maximum likelihood estimation, making it hard for model generalisation. We will exam the limitations of ICA in future simulations.

\subsection{Sparse representations of signals}
Here we introduces the idea of sparse signal processing and how it helps to solve underdetermined linear systems (dictionary). First we need to articulate the expression from the underdetermined system mentioned in previous parts. In BSS, `underdetermined system' means linear combination of \textbf{source signals} whereas here we refer to further linearly decompose the source signals into a given \textbf{dictionary}. In a more plain language, less number of examples than the data dimensionality involved are available for learning the dictionary. Sparse signal processing has numerous applications in compress sensing, image deniosing and super-resulation reconstrcution.

\subsubsection{Overcomplete dictionary}
\label{over_dict}
The key idea of sparse signal representation is to assume
that the sources are sparse, or can be decomposed into the
combination of a small number of signal components. By
sparse, we mean that most values in the signal or its transformed coefficients are zero. These signal components are called atoms, and the collection of all the atoms is referred to as a dictionary \cite{Mallat_Zhang1993}. In the general sparse representation framework, we can model a signal $y \in R^N$ as the linear combination of $D$ elementary signal atoms in dictionary $\mathbf{\Phi}$, such that.

\begin{equation}
    y = \alpha \mathbf{\Phi}
    \label{dict_qe1}
\end{equation}
where $\alpha$ is called the representation coefficients of $y$ in the dictionary $\mathbf{\Phi}$
%= \{\Phi_1, \cdots, \Phi_D \}
(the $N \times D$ matrix). 
In the case of overcomplete representations, the number of waveforms or atoms ($\varphi_i$) is higher than the dimension of the space in which $y$ lies, that is $D > N$. 

\begin{figure}[!htbp]
\centering
\includegraphics[width=0.45\textwidth]{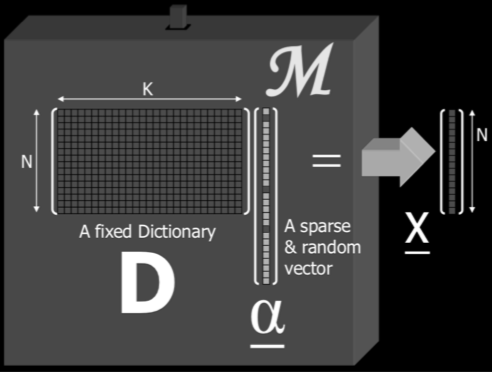}
\caption{Illustration of sparse representation using overcomplete dictionary}
\label{dic1}
\end{figure}

\subsubsection{Sparse decomposition}
\label{BSS_sparse_decomp}
The decomposition problem of a signal or image in predefined $\mathbf{\Phi}$ amounts to recovering the coefficient vector $\alpha$ in Equation (\ref{dict_qe1}). When $\mathbf{\Phi}$ is overcomplete, the solution is generally not unique. In that case, our goal is to recover the sparsest solution $\alpha$ which requires solving:
\begin{equation}
    \min_{\alpha}||\alpha||_0 \quad \text{s.t.} \quad y = \alpha \mathbf{\Phi}
\end{equation}

However, the above equation leads to an NP-heard optimisation problem due to its non-convextivity. Alternatively, we convexify the constraint by substituting the convex $\ell_1$-norm with the $\ell_0$-norm, leading to the following equation:
\begin{equation}
    \min_{\alpha} ||\alpha||_1 \quad \text{s.t.} \quad y = \alpha \mathbf{\Phi} 
    \label{l1_sparse}
\end{equation}
or\\
\begin{equation}
    \min_{\alpha}||\alpha||_1 \quad \text{s.t.} \quad || y - \alpha \mathbf{\Phi}||_2 \leq \sigma
\end{equation}

Such relaxations have led to a wide range of algorithms for signal reconstruction: the \textit{Basis Pursuit} based on linear programming  \cite{BPAtomicDcomp}, the greedy algorithm such as \textit{Matching Pursuit} \cite{Mallat_Zhang1993} and subspace techniques such as \textit{Subspace Pursuit} \cite{WdaiSP} and \textit{CoSaMP} (Compressive
Sampling Matching Pursuit) \cite{CoSaMP2008}. In the next section, we provide essential insights into the use of sparsity in BSS and we highlight the central role played by morphological diversity as a way of contrast between the sources.

\subsection{Morphological component analysis (MCA)}
We now introduce a practical algorithm named the morphological component analysis (MCA) aiming at decomposing signals in overcomplete dictionaries composed of a union of orthonormal basis, i.e. our dictionary is a concatenation of sub-dictionaries. In MCA setting, $y$ is a linear combination of $D$ morphological components. 
\begin{equation}
    y = \sum_{i=1}^D  \alpha_i \mathbf{\Phi}_i = \sum_{i=1}^D \varphi_i
    \label{Eq_dictionary}
\end{equation}
where $\mathbf{\{\Phi}_i\}$ are orthonormal basis whose columns are the atoms and is general normalised to a unit $\ell_2$-norm. Morphological diversity then relies on the incoherence between the subd-ictionaries. In terms of $\ell_0$-norm, this morphological diversity can be formulated as follows:
\begin{equation}
    \forall\{i,j\} \in \{1,\cdots,D\};\quad j \neq i \Rightarrow \; \lVert \varphi_i \Phi_i^T \rVert_0 < \lVert \varphi_j \Phi_i^T \rVert_0
\end{equation}
Intuitively, we can always find a sub-dictionary that one morphological component is highly sparse in it whereas other components are not very sparse. We therefore estimate the morphological components $\{\varphi_i\}_{i=1,\cdots,D}$ by solving the following convex minimisation problem.
\begin{equation}
    \{\varphi_i\} = \text{Arg} \: \min_{\{\varphi_i\}}\sum_{i=1}^D \lVert\varphi_i \mathbf{\Phi}_i^T \rVert_{1} + k \:\lVert y-\sum_{i=1}^D\varphi_i \rVert^2_2 +  \sum_{i=1}^D \gamma_i C_i (\varphi_i)
    \label{MCAequation}
\end{equation}
where $C_i$ implements constraints (e.g. TV correction) on component $\varphi_i$. Note that this minimisation problem differs from the problem setting in Equation (\ref{l1_sparse}) by relaxing the equality constraints to the later punishment term. A fast numerical solver called the \textit{Block-Coordinate Relaxation Method} \cite{BlockCoordinateMethod} was proposed to solve this kind of optimisation problem. 
The algorithm is given as follows:

\begin{algorithm}
\caption{The numerical algorithm for MCA} 
\label{algFramwork1} 
\begin{algorithmic} %这个1 表示每一行都显示数字
\REQUIRE ~~\\%Input
The sources $y$, dictionary $\Phi$, number of morphological components $D$, number of iterations $L_{max}$ and threshold $\delta$
\ENSURE ~~\\ %Output
Each morphological components $S_k$\\
\STATE Initialize $L_{max}$; number of iterations; threshold $ \delta= k L_{max}$;
\STATE 1. Perform $J$ times:
\STATE \qquad 2. Perform $D$ times:
\STATE \qquad \quad Update of $\varphi_i$ assuming all $\varphi_l$, $l \neq i$ are fixed:
\STATE \qquad \quad - Calculate the residual $r = \varphi - \sum_{l=1,l \neq i}^D \varphi_l$
\STATE \qquad \quad - Calculate the transform $\mathbf{\Phi^T}$ of $\varphi_i + r$and obtain $a_i = \mathbf{\Phi_i^T} (\varphi_i + r)$
\STATE \qquad \quad - Soft threshold the coefficient $a_i$ with the $\delta$ threshold and obtain $\Hat{a}_i$.
\STATE \qquad \quad - Reconstruct $\varphi_i$ by $\varphi_i = \mathbf{\Phi}_i \Hat{a}_i$
\STATE \qquad \quad - Apply the constraint correction $\varphi_i = \varphi_i - \mu \gamma_k \frac{\partial C_i}{\partial s_i}$
\STATE \qquad \quad - The parameter $\mu$ is chosen either by a line-search minimizing the overall 
\STATE 3. Update the threshold by $\delta = \delta - k$.
\STATE 4. If $\delta  > k$, return to Step 2. Else, finish.
\end{algorithmic}
\end{algorithm}

\subsection{Mutichannel morphological component analysis (MMCA)}
\begin{figure}[!htbp]
\centering
\includegraphics[width=0.7\textwidth]{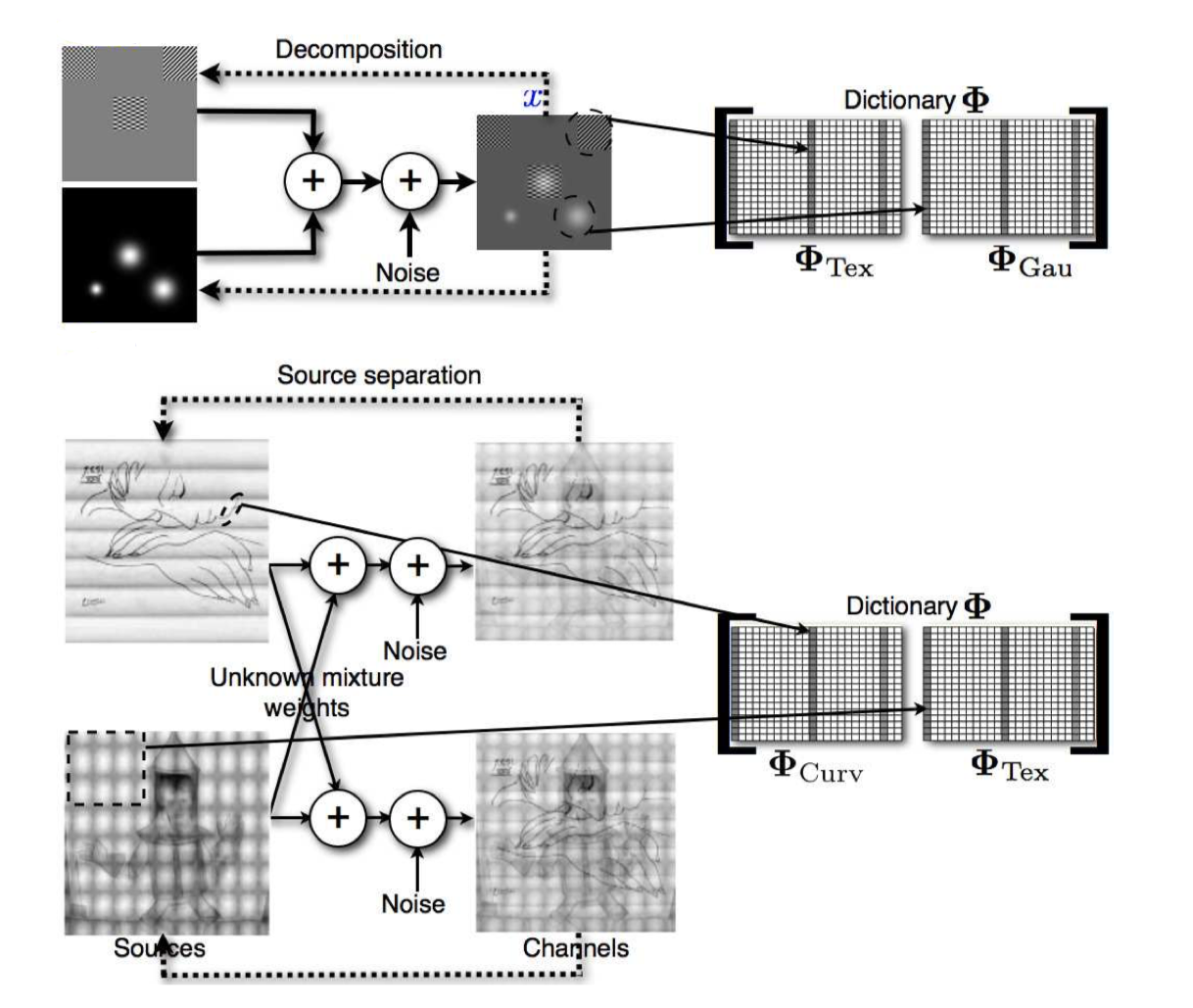}
\caption{Illustration of the image decomposition(top) using MCA and the blind source separation (bottom) using MMCA. For the bottom part, each source is itself a mixture of morphological components(texture and cartoon)}
\label{illu_mca_and_gmca}
\end{figure}

In this section, we extend the monochannel sparse decomposition problem described and characterized in last section to multichannel data. In the MMCA setting, we assumed that the sources $\mathbf{S}$ in Equation (\ref{mixing}) have strictly different morphologies (i.e. each source $s_i$ is assumed to be strictly sparsely represented in one particular orthonormal basis $\mathbf{\Phi}_i$) \cite{BobinJ_2007SaMD}. Figure \ref{illu_mca_and_gmca} precisely reveals the difference between MCA and MMCA. The top observation can be strictly divided in to texture and gaussian parts whereas the bottom multichannel observations are complex combinations of curvelets and texure components.\\

An iterative thresholding Block-Coordinate Relaxation algorithm similar to Algorithm (\ref{algFramwork1}) was proposed to solve the joint optimisation problem below.
\begin{equation}
    \{\mathbf{\tilde{A},\tilde{S}}\} = \text{Arg} \: \min_{\mathbf{A},\mathbf{S}}\sum_{k=1}^N \lVert s_k \mathbf{\Phi}_k^T \rVert_{1} + \lambda \:\lVert \mathbf{X} - \mathbf{AS} \rVert^2_2
    \label{MMCAequation}
\end{equation}
The equation above is very similar to Equation (\ref{MCAequation}) in MCA. Unfortunately, this MMCA criterion suffers from several drawbacks and particularly from an indeterminacy attached to the model structure \cite{BobinJ_2007SaMD}. For instance, the minimisation of this criterion may result in trivial solutions : $\mathbf{A} = \rho \mathbf{A}$ and $\mathbf{S} = \frac{1}{\rho} \mathbf{S}$ (the sparsity term can be minimised as desired as long as $\rho$ tends to $+\infty$) \cite{BobinJ_2007SaMD}. We therefore normalise the columns in matrix $\mathbf{A}$ at each iteration, as mentioned in Chapter \ref{ambiguitiesBSS}.\\

Let's introduce the $k^{\text{th}}$ channel residual $D_k = X - \sum_{j \neq k}a^j s_{j}$ (accounts for the part of that data unexplained by the other couples $\{a^j, s_j\}j \neq k$), the minimisation of the whole criterion (\ref{MMCAequation}) is equivalent to this joint minimisation problem.
\begin{equation}
    \{\tilde{s}_k, \tilde{a}^k\} = \text{Arg} \: \min_{\mathbf{A},\mathbf{S}} \lVert s_k \mathbf{\Phi}_k^T \rVert_{1} + \lambda \:\lVert D_k - a^ks_k \rVert^2_2
    \label{MMCAequation2}
\end{equation}
If we also assume the noise covariance matrix $\Gamma_n$ is known, the criterion new becomes:
\begin{equation}
    \{\mathbf{\tilde{A},\tilde{S}}\} = \text{Arg} \: \min_{\mathbf{A},\mathbf{S}} \lVert s_k \mathbf{\Phi}_k^T \rVert_{1} + Trace \{ (D_k - a^ks_k)\Gamma^{-1}(D_k - a^ks_k)^T\}
    \label{MMCAequation3}
\end{equation}
Zero the garident with respect to $s_k$ and $a^k$ leads to the following coupled equations
\begin{equation}
    \begin{cases}
       s_k = \frac{1}{a^k\Gamma^{-1}a^k}({a^k}^T - \frac{1}{2\lambda_k} Sign(s_k \mathbf{\Phi}_k) \mathbf{\Phi}_k^T)\\
       a^k = \frac{1}{s_ks_k^T}D_k s_k^T
    \end{cases}
\end{equation}
We then use the soft thresholding algorithm to solve for approximation of $s_k$ and $a_k$. Setting the threshold $\delta = \frac{\lambda_k}{2\lVert a^k\Gamma^{-1}a^k \rVert}$. Then considering a fixed $s_k$, the update on $a_k$ follows a simple least square linear regression.\\

\begin{algorithm}[!htbp] 
\caption{ The numerical algorithm for MMCA} 
\label{alg:Framwork} 
\begin{algorithmic}
\REQUIRE ~~\\%Input
The sources $S$, dictionary $\Phi$, number of morphological components $N$, number of iterations $L_{max}$ and threshold $\delta = k \cdot L_{max}$
\ENSURE ~~\\ %Output
Estimation $\hat{\mathbf{A}}$ and $\hat{\mathbf{S}}$
\STATE 1. Perform J times:
\STATE \qquad 2. Perform N times:
\STATE \qquad \quad - Normalisation and propagation of $a^k, s_k, \delta_k$ for scale invariance:\\
\STATE \qquad \quad - Estimation of $s_k$ assuming all $s_l$ , $l\neq k$ and $a_l$ are fixed
\STATE \qquad \quad - Calculate the residual $D_k = X - \sum_{l=1,l\neq k} a^ls_l$
\STATE \qquad \quad - Projection the residual $\hat{s}_k = \frac{1}{a^{kT}\Gamma_n^{-1}a^k}\Gamma_n^{-1}D_k$
\STATE \qquad \quad - Calculate $a_k = \hat{s}_k \mathbf{\Phi^T}_k$
\STATE \qquad \quad - Soft-thresholding the coefficients $a_k$ with the $\delta_k$ threshold and obtain $\hat{a}_k$
\STATE \qquad \quad - Reconstruct $s_k$ by $s_k = \hat{a}_k \mathbf{\Phi}_k$
\STATE \qquad \quad - Estimation of $a_k$ by $a_k$ assuming all $s_l$ and $a^l_{l \neq k}$ are fixed $a_k = \frac{1}{s_k s_k^T D_ks_k^T}$
\STATE 3. Update the threshold by $\delta = \delta - k$.
\STATE 4. If $\delta  > k$, return to Step 2. Else, finish.
\end{algorithmic}
\end{algorithm}

\subsection{Generalised morphological component analysis (GMCA)}
We now apply the idea of morphological diversity to more generalised blind source separation problems. In GMCA, we assume each source is modelled as a weighted sum of $D$ morphological components where each component is sparsely represented in a specific basis.
GMCA pursuits an unmixing scheme, through the estimation of $\mathbf{A}$, which leads to the sparsest sources $\mathbf{S}$ in the dictionary $\mathcal{D}$. This is expressed in a Lagrangian form:

\begin{equation}
    \{\mathbf{\tilde{A},\tilde{S}}\} = \text{Arg} \: \min_{\{\mathbf{A,S}\}} \sum_{i=1}^n \sum_{k=1}^D \lVert\varphi_{ik} \mathbf{\Phi}_k^T \rVert_{1} + \lambda \:\lVert \mathbf{X} - \mathbf{AS} \rVert^2_2
    \label{GMCAequation}
\end{equation}

The product $\mathbf{AS}$ can be split into $n\times D$ multichannel morphological components: $AS = \sum_{i,k}a^i\varphi_ik$. Based on this decomposition, an alternating minimisation algorithm was proposed to estimate iteratively one term at a time \cite{BobinJ_2007SaMD}. Again, each column of $\mathbf{A}$ is forced to have unit norm at each iteration to avoid the classical scale indeterminacy of the product in (\ref{GMCAequation}). Define the multichannel residual by $\mathbf{X}_{i,k} = \mathbf{X} - \sum_{\{p,q\}\neq \{i,k\}} \alpha^p \varphi_{pq}$ as part of the data unexplained by the multichannel morphological component $\alpha^i \varphi_{ik}$. Estimating the morphological component $\varphi_{ik} = \alpha_{ik}\mathbf{\Phi}_k$ assuming $\mathbf{A}$ and $\varphi_{\{pq\} \neq \{ik\}}$ are fixed leads to: 
\begin{equation}
    \tilde{\varphi}_{ik} = \text{arg} \; \min_{\{\varphi_{ik}\}} \lVert \varphi_{ik}\mathbf{\Phi}_k^T\rVert_{1} + \lambda \; \lVert \mathbf{X}_{i,k} - a^i\varphi_{ik}\rVert_2^2
\end{equation}
Similarly to MMCA, GMCA uses the component-wise iterative thresholding algorithm which is summarized as follows. Note the difference notations of morphological components coefficients $\alpha_{ik}$ and mixing matrix coefficients $a^i$.\\

\begin{algorithm}[!htbp] 
\caption{ The numerical algorithm for GMCA.} 
\label{algFramwork3} 
\begin{algorithmic}
\REQUIRE ~~\\%Input
The sources $S$, dictionary $\Phi$, number of morphological components $D$, number of iterations $L_{max}$ and threshold $\delta = k \cdot L_{max}$
\ENSURE ~~\\ %Output
Estimation $\tilde{\mathbf{A}}$ and $\tilde{\mathbf{S}}$
\STATE 1. Perform n times:
\STATE \qquad 2. Perform D times:
\STATE \qquad \quad - Compute the residual term $r_{ik}$ assuming the current estimates $\tilde{\varphi}_{\{pq\} \neq ik}$ are fixed; 
\STATE \qquad \quad $r_{ik} = \tilde{a}^{iT}(\mathbf{X} - \sum_{\{p,q\} \neq \{i,k\}} \tilde{a}^p \tilde{\varphi}_{pq}) $
\STATE \qquad \quad - Estimate the current coefficients of $\tilde{\varphi}_{ik}$ by thresholding with threshold $\delta$
\STATE \qquad \quad $\tilde{\alpha}_{ik} = \lambda_{\delta}(r_{ik}\mathbf{\Phi}_k^T)$
\STATE \qquad \quad - Reconstruct $\varphi_{ik}$ by $\varphi_{ik} = \tilde{\alpha}_{ik} \mathbf{\Phi}_k$
\STATE \qquad \quad - Estimation of $a_k$ by assuming all $\varphi_{pq}$ and $a^{p \neq k}$ are fixed 
\STATE \qquad \quad $\tilde{a}^i = \frac{1}{\tilde{s_i}^2}(\mathbf{X} - \sum_{p\neq i}^n \tilde{a}^p\tilde{s}_p)\tilde{s}_i^T$
\STATE 3. Update the threshold by $\delta = \delta - \lambda$.
\STATE 4. If $\delta>k $, return to Step 2. Else, finish.
\end{algorithmic}
\end{algorithm}

\subsection{FastGMCA algorithm (FGMCA)}
In the last section, we described GMCA algorithm which needs the projection of the residual into the dictionary space at each iteration $\tilde{\alpha}_{ik} = \lambda_{\delta}(r_{ik}\mathbf{\Phi}_k^T)$. Note that the application of $\mathbf{\Phi}_k^T$ will consume most of the computation power. Thus, GMCA could be very computationally demanding for large scale, high dimensional problems \cite{BobinJ_2007SaMD}. In practice, we apply the an improved version coined fast GMCA by adding some assumptions to the original problem.\\

We assume each row of $\mathbf{\Theta_X}=\mathbf{X}D^T$ stores the decomposition of each observed channels in $D$. And each row of $\mathbf{\Theta_S}=\mathbf{S}D^T$ stores the decomposition of each source. 
If the supports ($\ell_0$ decompostion) of $\mathbf{X}$ and $\mathbf{S}$ satisfies 
\begin{equation}
    \Delta_D(x_i) = \sum_{j=1}^{n}\alpha_{ij}\Delta(s_j)
\end{equation}
Then we can rewrite the Lagrangian form as follows.

\begin{equation}
    \{\mathbf{\tilde{A},\tilde{S}}\} = \text{Arg} \: 
    \lVert \mathbf{\Theta_S} \rVert_0 + \lambda \:\lVert \mathbf{\Theta_X} - \mathbf{A\Theta_S} \rVert^2_2
\label{fast-gmca}
\end{equation}

To conclude, the fast GMCA algorithm works in the sparse transformed domain and omits the dictionary decomposition process at each iteration. Thus we can precompute the projection of the mixtures in, for example the wavelet domain and run the FastGMCA algorithm in that domain. This will significantly accelerate our blind separation process. Now write fast GMCA in a stepwise flavour.

\begin{algorithm}[!htbp] 
\caption{The numerical algorithm for FastGMCA}
\label{alg:Framwork} 
\begin{algorithmic}
\REQUIRE ~~\\%Input
The obervations $\mathbf{Y}$, dictionary $\mathbf{\Phi}$, number of morphological components $N$, number of iterations $L_{max}$ and threshold $\delta^{(0)} = k \cdot L_{max}$
\ENSURE ~~\\ %Output
Estimation $\tilde{\mathbf{A}}$ and $\tilde{\mathbf{S}}$

\STATE 1. Decompose the mixture in transformed domiain an obtain $\mathbf{\Theta_X}$\\

\STATE 2. While each $\delta$ is higher than a given lower threshold $\delta^{(0)}$:
\STATE \quad - Update $\mathbf{\Theta_S}$ with thresholding operator $\lambda_{\delta}$ and $\mathbf{A}$ is fixed at $h^{th}$ iteration.\\
\quad \quad $\mathbf{\hat{\Theta}}^{(h+1)}_S = \lambda_{\delta} (A^{\dagger^{(h)}}\mathbf{\Theta_X})$
\STATE \quad - Update $\mathbf{A}$ by a least-square estimate assuming $\mathbf{\Theta_S}$ is fixed.\\
\quad \quad $\mathbf{\hat{A}}^{(h+1)} = \mathbf{\Theta_X} \mathbf{\hat{\Theta}_S}^{(h)^T} 
\mathbf{(\hat{\Theta}_S}^{(h)} \mathbf{\hat{\Theta}_S}^{(h)^T})^{-1}$
\STATE \quad - Decrease $\delta$.
\end{algorithmic}
\end{algorithm}

\subsection{Blind source separation based on adaptive dictionary learning}
In previous sessions, we mentioned about sparse decomposition using prescribed overcomplete dictionaries such as Wavelet, Curvelet or unions of orthonormal transform bases. This method works well when the original sources have components that are largely different from each other in the transform domain. However, this may not lead to the most sparsified decomposition of each individual sources, or in another word, the dictionaries found may not be appropriate in the sense that they may fit better the mixtures rather than the sources.\\

As described before, dictionary learning has wide applications in compress sensing and image denoising. Apart from the pursuit algorithms described in \ref{over_dict} that finds the sparse coefficients with respect to a given dictionary. Recent research is concentrated on obtaining a adapting dictionary in order to achieve best sparse signal representations. Michal Aharon designed the K-SVD \cite{AharonM2006KAaf} algorithm which uses a K-means clustering like algorithm and columnwise updating flavor of the dictionary atoms. Dai and Tao proposed the SimCo method \cite{6340354} based on K-SVD which avoids falling into a singular point during the optimisation process. It is worthy noting that these methods only gives an appropriate solution to the dictionary learning problem formulated in in Equ.(\ref{equ14}), but finding a global minima (for sure) is still an open question.\\
\begin{equation}
    \min_{\alpha}||\alpha||_1 \quad \text{s.t.} \quad || \mathbf{Y} - \mathbf{X} \mathbf{\Phi}||^2 \leq \sigma
    \label{equ14}
\end{equation}
Motivated by the idea of image denoising we can adapt MMCA to learned local dictionaries from the mixed sources within the separation process\cite{VAbolghasemi2012}. Hence we does not need any prior knowledge about the sparse domain of the sources. We now introduce the K-SVD dictionary learning algorithm which will be applied later in experiment section.\\

\subsubsection{K-SVD dictionary learning}
% The K-SVD algorithm involves two basic steps, which together constitute the algorithm iteration: (i) the signals in X are sparse-coded given the current dictionary estimate, producing the sparse representations matrix Γ, and (ii) the dictionary atoms are updated given the current sparse representations; see Algorithm 3. The sparse-coding part (line 5) is commonly implemented using OMP. The dictionary update (lines 6-13) is performed one atom at a time, optimizing the target function for each atom individually while keeping the rest fixed.\\
The K-SVD algorithm can be decomposed into two stages, which are executed alternatively as what we did in the iterative relaxation method. First we can use any pursuit algorithm (BP, OMP) to calculate the sparse coefficients $\mathbf{X}$. This is called the sparse coding stage. Then we proceed to the codeword update stage. In Equation (\ref{equ14}), assuming that both $\mathbf{\Phi}$ and $\mathbf{X}$ is fixed and we only consider one atom $\phi_k$ in dictionary $\mathbf{\Phi}$ and the coefficients $x^k$ ($k^{th}$ row in $\mathbf{X}$) corresponding to it, the penalty term in the objective function can be rewritten as

\begin{equation}
\begin{split}
    ||\mathbf{Y} - \mathbf{\Phi X} ||^2 & = || \mathbf{Y} - \sum_{j=1}^K \phi_j x^j_T||^2\\
    & = ||(\mathbf{Y} - \sum_{j\neq k} \phi_j x^j_T ) - \phi_k x_T^k||^2\\
    & = || \mathbf{E}_k - d_k x_T^k ||^2
\end{split}
\end{equation}
where $\mathbf{E}_k$ stands for the residual. 
In order to force the sparsity in $\mathbf{X}$, we need to extract the columns in $\mathbf{E}_k$ correspond to non-zero elements in $x^k$. That is,

\begin{equation}
\begin{aligned}
    w_k &= \left\{ i \;\middle|\; 1 \leq i \leq K,\; x^k(i) \neq 0 \right\} \\ 
    \Omega_k &= \text{concatenation of } N \times w_k \\
    \mathbf{E}_k &= \Omega_k \mathbf{E}_k
\end{aligned}
\end{equation}

$\Omega_k$ is a mask consisting 0 and 1s so that we only consider atoms that refer to a certain row in the sparse coefficients. Hence we have split the term $\mathbf{\Phi X}$ to $k$ rank-1 matrices. Among those, only the $k^{th}$ atom remains in the question. This is a least square problem that can be directly solved using singular value decomposition (SVD). 
\begin{equation}
    \mathbf{E}_k = U\Sigma V^T
\end{equation}
We define the solution for $\phi_k$ as the first column of $U$ and the coefficient vector $x^k$ as the first column of $V$. In the same manner, K-SVD sweeps through all columns always use the most updated coefficients as they emerge from preceding SVD steps. To conclude, The K-SVD algorithm obtains the dictionary update by $K$ separate SVD computations, which explains its name.\\

\subsubsection{K-SVD+MMCA}
The combination of K-SVD and MMCA in blind source separation 
is similar to the conventional MMCA algorithm apart from it requires updating of three matrices, the estimated mixing $\mathbf{A}$, the estimated source $\mathbf{S}$ and the dictionary $\mathbf{\Phi}$. A stepwise algorithm is displayed below

\begin{algorithm}[!htbp] 
\caption{The numerical algorithm for K-SVD+MMCA} 
\label{alg:Framwork} 
\begin{algorithmic}
\REQUIRE ~~\\%Input
The sources $S$, dictionary $\Phi$, number of morphological components $N$, number of iterations $L_{max}$ and threshold $\delta^{(0)} = k \cdot L_{max}$.
\ENSURE ~~\\ %Output
Estimation $\hat{\mathbf{A}}$ and $\hat{\mathbf{S}}$.

\STATE 1. Initialse $\mathbf{\Phi}$ to a known overcomplete dictionary.\\
\STATE 2. Set $\mathbf{A}$ to a radom column-normalised matrix.\\
\STATE 3. $\mathbf{X} = A^T Y$.\\

\STATE 4. for $L_{max}$ iterations:\\
\quad \quad for $j = 1:N$:\\
\quad \quad \quad - Extract patches from $x_j$.\\
\quad \quad \quad - Update coefficient $a_j$ using OMP.\\
\quad \quad \quad - Update $\Phi_j$ using K-SVD.\\
\quad \quad \quad - Calculate the residual $\mathbf{E_j} = \mathbf{Y} - \sum_{l\neq j}a_lx_l^T$. \\
\quad \quad \quad - Compute $x_j$. \\
\quad \quad \quad - $a_j = \mathbf{E_j}x_j$ \\
\quad \quad \quad - Normalise $a_j$  \\
\STATE 5. Decrease $\sigma$ until stopping criterion is met. \\
\end{algorithmic}
\end{algorithm}

\section{Block Sparse K-SVD algorithm applied to BSS}
Last section proves the strength of sparsity based methods applied to blind source separation problem. Inspired by using adaptively learned local dictionary in MMCA/GMCA, we aim to create a new adaptive dictionary learning algorithm combined with BSS. The idea driven behind is simple that, certainly will we acquire a better separation result if we improve the level of sparsity of the dictionary,.\\

A variety of sparse dictionary learnging algorithm have been proposed in the literature for this purpose, based on the K-SVD algorithm. The block-sparse dictionary learning algorithm is proposed by Lihi in \cite{dictionary_block_sparse}. It exploits the hidden structure that is intrinsic in the signals for producing more efficient sparse representations. In the following content, we introduce the theory of this algorithm and try to embed it in the blind source separation process. The algorithm consists of two steps: a block structure update step (SAC) and a dictionary update step (BK-SVD). \\

\subsection{Problem definition}
\begin{figure}[!htbp]
\centering
\includegraphics[width=0.7\textwidth]{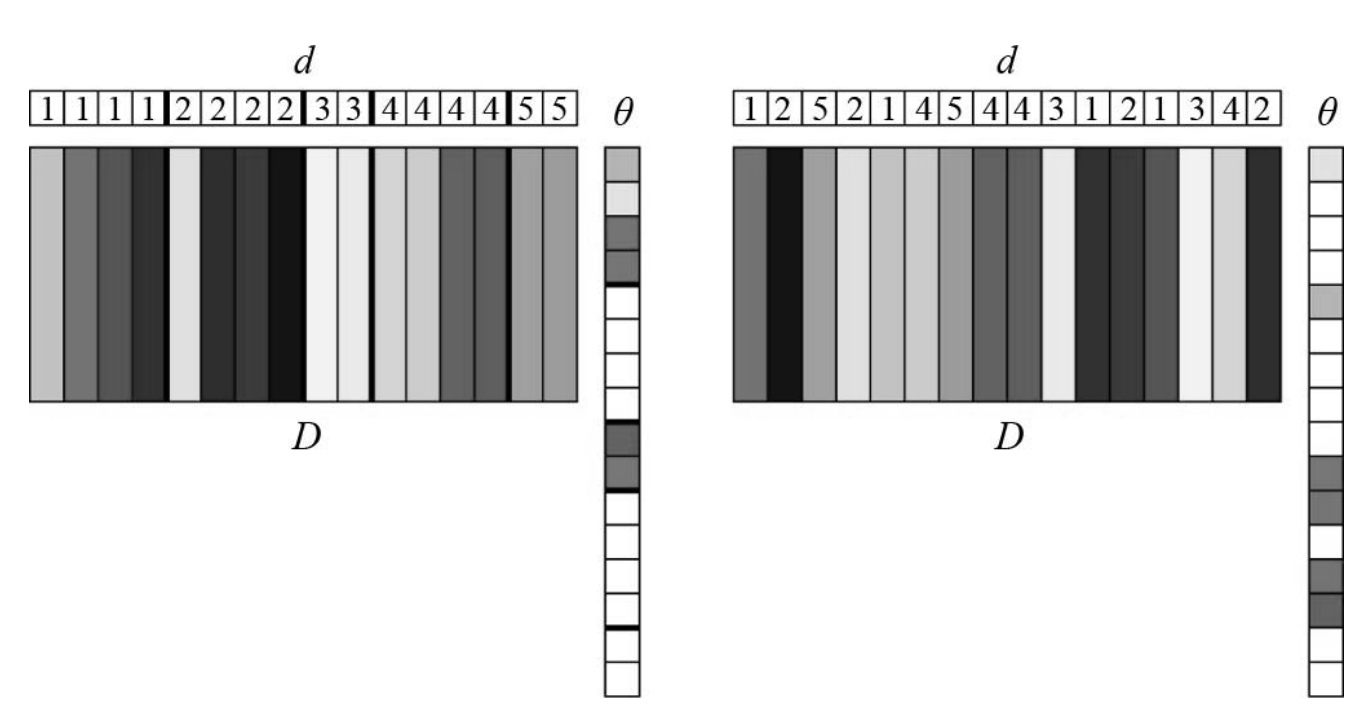}
\caption{Two equivalent representations with different block structure.}
\label{block_dict_compare}
\end{figure}

Given a set of signals $\mathbf{Y} = \{y_i\} \in R^N$, we wish to find an overcomplete dictionary $\mathbf{\Phi}$ in $R^{N \times K}$ whose atoms are sorted in blocks, correspondingly the non-zero coefficients representations $\mathbf{X} = \{x_i\}$ are concentrated in a fixed number of blocks. More specifically, suppose dictionary atoms sorted in blocks that enable \textbf{block-sparse} representations of input signals. Each block has its own label, indexed as $d_i$. We claim that a vector $\mathbf{X}\in R^K$ is $k$-block-sparse over $d$ if its non-zero entries under certain block strucutre is less than $k$. This is denoted by 
\begin{equation}
    ||x||_{0,d} = k
\end{equation}
which means there are $k$ number of non-zero blocks having block structure $d$. Figure \ref{block_dict_compare} presents two equivalent examples, The dictionary can be expressed as 5 blocks but with 2-block-sparse presentations.\\

We can now formulate the problem is a mathematical way that we intent to find a dictionary $\mathbf{\Phi}$ and a block structure d, with maximal block size s, that lead to optimal k-block sparse representations $\mathbf{X} = \{x_i\}^L_i $ for signals in $\mathbf{Y}$. The objective function is given as below.\\
\begin{equation}
\begin{aligned} 
    \min_{D,d,\mathbf{X}} ||\mathbf{Y} -\mathbf{\Phi}\mathbf{X}||\\
    \text{s.t.} \quad ||x_i||_{0,d} \leq k, \; i = 1,...L\\
    |d_j| \leq s, \; j = 1,....B
    \label{bskvsd_1}
\end{aligned}
\end{equation}

where $d_j$ is the set of indices belonging to block $j$, $s$ is the maximum block size and $B$ is
is the total number of blocks (number of dictionary columns divided by the block size). When the maximal block size is set to 1, the proposed algorithm reduces to normal K-SVD.\\

\subsection{Algorithm preview}
Like other optimisation problems in previous sections, the problem in Equation (\ref{bskvsd_1}) is non-convex. We therefore adopt the coordinate relaxation technique. The initial dictionary can be set up as a DCT dictionary or any random collection of $K$ signals. Then the block structure is solved by the \textit{sparse aggolmerative clustering} (SAC) algorithm\cite{Johnson1967}. Agglomerative clustering is a `bottom-up' approach who groups according to distance metric (i.e. $\ell_0$ norm). SAC algorithm solves for an optimal block structure refer to input command ($k$ and $s$) while keeping the dictionary fixed.

\begin{equation}
\begin{aligned}
    \left[d^{(m)}, \;\mathbf{\Phi}^{(m)}\right] = \text{arg} \; \min_{X,d} \quad || \mathbf{Y} - \mathbf{\Phi}^{(m-1)}\mathbf{X}||_2 \\
    \text{s.t.} \quad ||x_i||_0,d \leq k, \quad i = 1,...,L\\
    |d_j| \leq s, \; j = 1,....B
\end{aligned}
\end{equation}

\begin{figure}[!htbp]
\centering
f\includegraphics[width=0.75\textwidth]{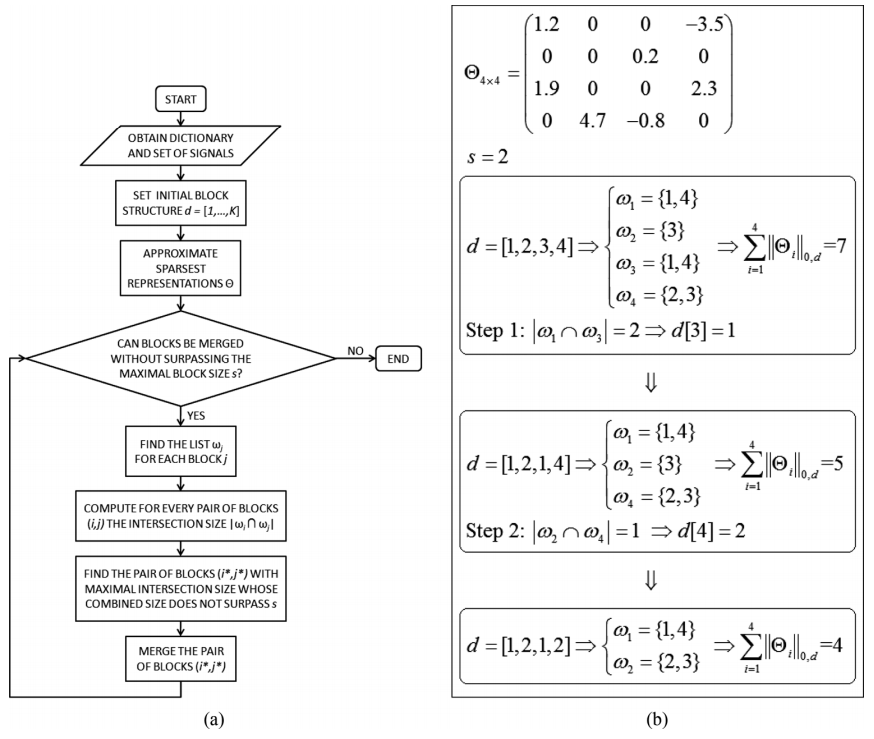}
\caption{(a) A flow chart about the SAC algorithm; (b) A stepwise example of the SAC algorithm.}
\label{flow_sac}
\end{figure}

Figure (\ref{flow_sac}) presents a detailed example of the clustering process of the SAC algorithm. In this example, the maximal block size is set to be 2. At the first iteration, row 1 and row 3 in the dictionary has the largest intersection, by which means their $\ell_0$ norm has the largest similarity ($\omega = \{1,4\}$). Consequently, they are merged together. In the next step, row 2 and row 4 have the largest intersection and are merged. Up to now, no acquisitions an be further executed because of the maximal block size is limited by 2. \\

One may wonder that the SAC algorithm only deals with the sparse coefficients but not the dictionary directly. This is because rows of the coefficients $\mathbf{X}$ exhibits a similar pattern on non-zeros as the columns of the dictionary block. In other words, grouping coefficients is equivalent to grouping the dictionary atoms according to the sparsity pattern. Furthermore, merging blocks is always beneficial to the reconstruction accuracy of one dictionary, due to the reconstruction result is a liner summation of atoms, grouping only rewrite the coefficients in $\mathbf{X}$, which does not deprave the final result.\\

After having successfully determined the optimal block structure. We then solve for new dictionary while keeping $\mathbf{\Phi}^{m}$ fixed. The objective function becomes.
\begin{equation}
\begin{aligned}
    \left[\mathbf{X}^{(m)}, \;\mathbf{\Phi}^{(m)}\right] = \text{arg} \; \min_{\Phi,X} \quad || \mathbf{Y} - \mathbf{\Phi}^{(m-1)}\mathbf{X}||_2 \\
    \text{s.t.} \quad ||x_i||_0,d \leq k, \quad i = 1,...,L
\end{aligned}
\label{BKSVD1}
\end{equation}
The author in \cite{dictionary_block_sparse} proposed block K-SVD (BK-SVD), a natural extension of the K-SVD algorithm to solve (\ref{BKSVD1}). BK-SVD algorithm employs a similar `columnwise' atom updating manner as in K-SVD but forces the learned dictionary to have higher sparsity level. We first fix $\mathbf{\Phi}^{m-1}$ and can use any pursuit method (e.g. Batch-OMP) to solve (\ref{BKSVD1}) which reduces to.

\begin{equation}
\begin{aligned}
    \mathbf{X}^{(m)} = \text{arg} \; \min_X \quad || \mathbf{Y} - \mathbf{\Phi}\mathbf{X}^{(m-1)}||_2 \\
    \text{s.t.} \quad ||x_i||_0,d \leq k, \quad i = 1,...,L
\end{aligned}
\end{equation}

Then, to obtain $\mathbf{\Phi}^{(m)}$, fix $\mathbf{X}^{(m)}$, $d$ and $\mathbf{Y}$. The calculation procedure is same as K-SVD, where the blocks of atoms (not standalone atoms) in the dictionary is updated sequentially, alongside with the corresponding nonzero coefficients in $\mathbf{X}^{(m)}$. The key difference between K-SVD and BK-SVD is that, in K-SVD only the highest rank component of the residual is updated, resulting in one atom change. Conversely in BK-SVD, atoms in the same block can be updated simultaneously.\\

We have now introduced the whole block sparse dictionary learning algorithm. Extend the idea of K-SVD+MMCA algorithm, we can replace the step of calculating sparse coefficients and dictionary atoms by the newly proposed SAC+BK-SVD algorithm. The overall BSS framework using SAC+BK-SVD is summarised below.

\begin{algorithm}[!htbp] 
\caption{The numerical algorithm for SAC+BK-SVD+MMCA} 
\label{alg:Framwork} 
\begin{algorithmic}
\REQUIRE ~~\\%Input
The sources $S$, dictionary $\Phi$, number of morphological components $N$, number of iterations $L_{max}$ and threshold $\delta^{(0)} = k \cdot L_{max}$.
\ENSURE ~~\\ %Output
Estimation $\hat{\mathbf{A}}$ and $\hat{\mathbf{S}}$.

\STATE 1. Initialse $\mathbf{\Phi}$ to a known overcomplete dictionary.\\
\STATE 2. Set $\mathbf{A}$ to a radom column-normalised matrix.\\
\STATE 3. $\mathbf{X} = \mathbf{A}^T\mathbf{Y}$.\\

\STATE 4. for $L_{max}$ iterations:\\
\quad \quad for $j = 1:N$:\\
\quad \quad \quad - Extract patches from $x_j$.\\
\quad \quad \quad - Fix dictionary $\mathbf{\Phi}^{(m-1)}$ and update coefficients $\alpha^{m}$ and block structure $d^{(m)}$ by applying sparse agglomerative clustering.\\
\quad \quad \quad- Fix the block structure $d^{(m)}$ and update the dictionary atoms $\mathbf{\Phi}^{(m)}$ by applying BK-SVD.\\
\quad \quad \quad - Calculate the residual $\mathbf{E_j} = \mathbf{Y} - \sum_{l\neq j}a_lx_l^T$. \\
\quad \quad \quad - Compute $x_j$. \\
\quad \quad \quad - Calculate mixing matrix column $a_j = \mathbf{E_j}x_j$ \\
\quad \quad \quad - Normalise $a_j$  \\
\STATE 5. Decrease $\sigma$ until stopping criterion is met. \\
\end{algorithmic}
\end{algorithm}

\subsection{Complexity analysis}
The standard algorithm for agglomerative clustering (SAC) has a time complexity of ${\displaystyle {\mathcal {O}}(K^{3})}$ (K is the number of dictionary atoms in this occasion). This makes it not suitable for even medium datasets. However the complexity of BK-SVD requires $s$ times less number of singular value computations than the K-SVD algorithm. Because of the simultaneous updating of atoms belong to the same block.
As proved in \cite{OMP_KSVD}, K-SVD algorithm has a complexity of $\displaystyle {\mathcal {O}}((ks)^2K + 2NK)$. Because of the block nature of BK-SVD, we expect BK-SVD to have a complexity of $\displaystyle {\mathcal {O}}((k)^2K + 2NK)$, where $k$ is the sparsity level (number of sparse blocks) and $s$ is the maximal block size. The total compexity of combined SAC+BK-SVD is therefore $\displaystyle {\mathcal {O}}((k)^2K + 2NK+K^3)$. Therefore the overall convergence rate of the proposed BSS framework will be slower than simple K-SVD method.\\

\section{Experiments on Image Source Separation}
\subsection{Software requirements}
The software in this project has prerequisite on several opensource Matlab Toolboxes (i.e. WaveLab 850, K-SVD, MCALab110 and image processing toolbox). Complete version of my code can be found on Github\footnote{https://github.com/Dieselmarble/FYP}. In addition, I wrote my own block-sparse BSS Toolbox which involves some novelty.

\subsection{Solve the BSS scale and permutation indeterminacy}
As mentioned in previous sections, BSS methods suffers from ambiguities that the estimated source can be scaled permuted up to any arbitrary order. The solution to this problem is equivalent to an optimal assignment algorithm. An intuitive method would be comparing the similarity metric (e.g. Euclidean distance) between every estimated source and the original sources. It is easy to see that the complexity is $O(n^2)$. Unfortunately, the optimal assignment may not, in general, be attained in this way \cite{1261953}. In this report, we employ the Kuhn-Munkres Algorithm (also well-known as the Hungarian Algorithm)
with a higher complexity $O(n^3)$ (n is the number of channels) but guarantees the optimal assignment to calculate the permutation matrix.

\subsection{Image decomposition}
\begin{figure}[!htbp]
\centering
\subfigure[]{
\begin{minipage}[b]{0.23\linewidth}
\includegraphics[width=1\linewidth]{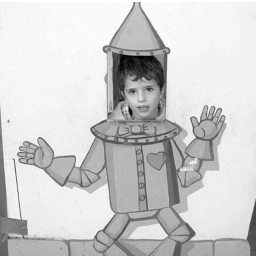}\vspace{1pt}
\includegraphics[width=1\linewidth]{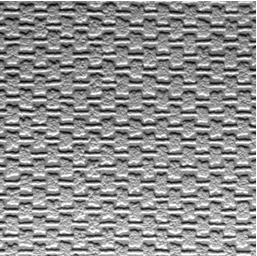}
\end{minipage}}
\subfigure[]{
\begin{minipage}[b]{0.23\linewidth}
\includegraphics[width=1\linewidth]{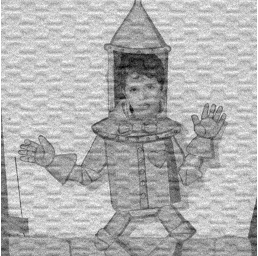}\vspace{1pt}
\includegraphics[width=1\linewidth]{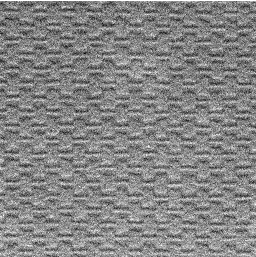}
\end{minipage}}
\subfigure[]{
\begin{minipage}[b]{0.23\linewidth}
\includegraphics[width=1\linewidth]{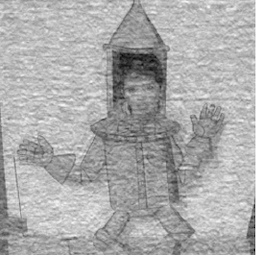}\vspace{1pt}
\includegraphics[width=1\linewidth]{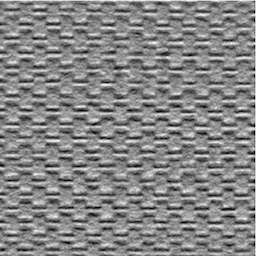}
\end{minipage}} 
\subfigure[]{
\begin{minipage}[b]{0.23\linewidth}
\includegraphics[width=1\linewidth]{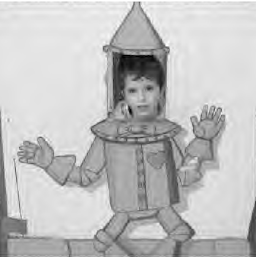}\vspace{1pt}
\includegraphics[width=1\linewidth]{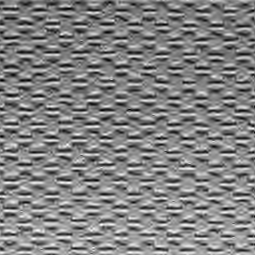}
\end{minipage}}
\caption{Experiment1: Image segmentation (PSNR = 20dB); \textbf{(a)}: Original sources; \textbf{(b)}: FastICA outputs; \textbf{(c)}: MCA outputs; \textbf{(d)}: MMCA outputs}
\label{segmentation_Im}
\end{figure}

In this section, we turn to use MMCA to separate two-dimensional data and compare the result with the standard ICA source separation techniques. In Figure (\ref{segmentation_Im}a) are two source signals, one of which is oscillating textures while another is  a `boy' image. Curvelet transform is selected as the dictionary for source 1 and discrete consine transform is selected for source 2. This is similar to the idea of a double sparse dictionary therefore we can decompose the mixtures into cartoon and texture. Figure (\ref{segmentation_Im}c) illustrates the separated image using MCA under the presence of 20dB Gaussian noise. It can be shown that MCA is able to split the texture and cartoon parts. However, the reconstruction quality of the `boy' image using curvelet dictionary in MCA does not give satisfactory result though. We think the decomposition presenting in the dictionary domain may not be extremely sparse, as some of the mixtures can still be seen in the output image. \\

\begin{figure}[!htbp]
\centering
\begin{minipage}[b]{0.49\textwidth}
\includegraphics[width=\textwidth]{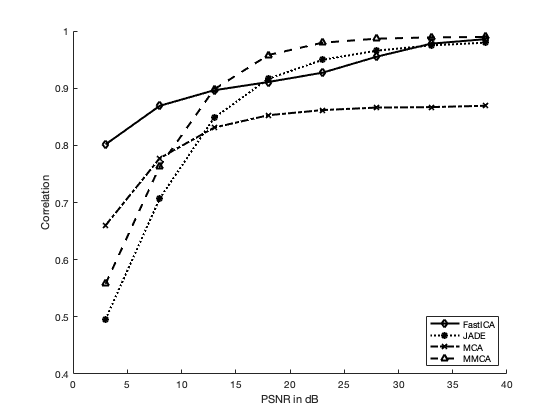}
\caption{Evolution of the correlation coefficient between original and estimated sources as the noise variance varies.}
\label{segmen_1}
\end{minipage}
\begin{minipage}[b]{0.49\textwidth}
\includegraphics[width=\textwidth]{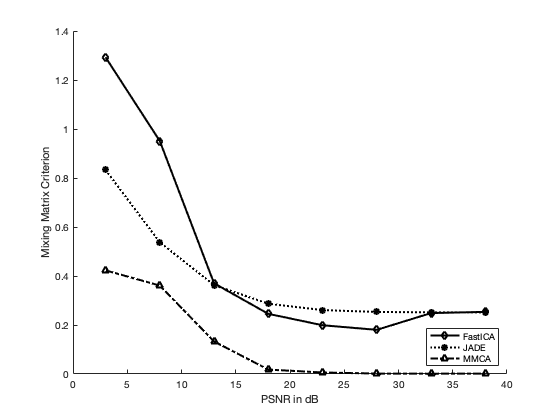}
\caption{Evolution of the mixing matrix criterion (after indeterminacy corrected) as the noise variance varies.}
\end{minipage}
\label{segmen_2}
\label{imapint1}
\end{figure}

Note that the MCA algorithm, unlike BSS methods, only takes one single combination of sources ($m = 1$). Now we extend the observed mixtures to a multichannel case and BSS techniques applies. The correlation coefficient between two sources are only 0.07. Hence the independence assumption for ICA methods is valid here. We create four mixtures from two source images.
Figure (\ref{segmentation_Im}d) shows that MMCA is clearly able to efficiently separate the original source images, achieving better visual results than FastICA in (b). Quantitatively, Figure \ref{segmen_1} shows the correlation between the original sources and those estimated. As the data noise variance increase, 
MMCA (dashed line) clearly achieves better estimation quality and shows clear robustness compared to non de-noised ICA methods. In addition, one can note that both JADE and FastICA provides similar performance.\\

Figure \ref{segmen_1} plots the matrix estimation error is defined as $||\mathbf{I_n} - P\tilde{\mathbf{A}}^{+}\mathbf{A} ||$, after elimination the effect of the permutation and scale indeterminacy. Contrasting with standard ICA methods, MMCA iteratively estimates the mixing matrix from coarse (i.e. smooth) versions of the sources and thus is not penalized by the presence of noise. As a consequence, MMCA is clearly more robust to noise than standard ICA methods, even under very noisy context \cite{BobinJ_2006Mdas}. This result reflects our expectation in section \ref{ica_defect} that ICA is not robust under the additive Gaussian noise setting.\\

The results in this experiment proves that sparsity based methods successfully handle the image segmentation task. Especially under the multichannel case, MMCA significantly outperforms the standard ICA methods in terms of separation quality and robustness.\\

% -------------------------
% -------------------------

\newpage
\subsection{Blind image source separation}
In this experiment, we mix 4 pictures into 10 channels. The source images are picked as they contains similar morphologies. Classical ICA methods and MMCA, GMCA are applied to separate the source. Unfortunately the FastICA methods is not able to separate the original source. We hence adopt a sophisticated variant of it abbreviated as EFICA. Moreover, it has been proved in \cite{BobinJ_2007SaMD} that using a single overcomplete DWT dictionary or a union of DCT and DWT dictionary in GMCA provides similar results. We therefore use a discrete wavelet dictionary in FastGMCA (FGMCA).\\

\begin{figure}[!htbp]
\centering
\subfigure[]{
\begin{minipage}[b]{0.17\linewidth}
\includegraphics[width=1\linewidth]{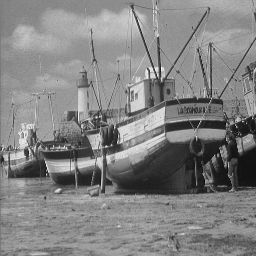}\vspace{4pt}
\includegraphics[width=1\linewidth]{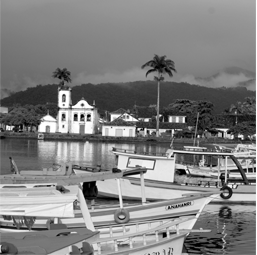}\vspace{4pt}
\includegraphics[width=1\linewidth]{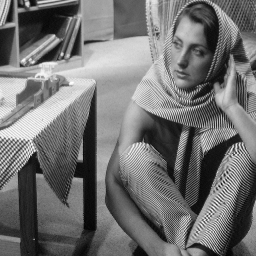}\vspace{4pt}
\includegraphics[width=1\linewidth]{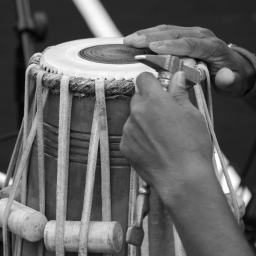}
\end{minipage}}
\subfigure[]{
\begin{minipage}[b]{0.17\linewidth}
\includegraphics[width=1\linewidth]{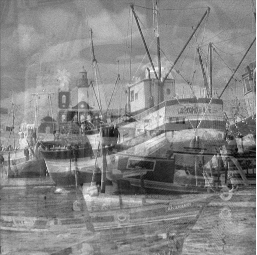}\vspace{4pt}
\includegraphics[width=1\linewidth]{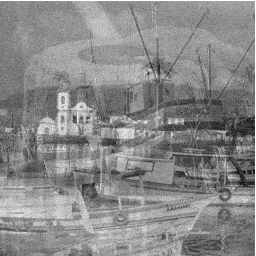}\vspace{4pt}
\includegraphics[width=1\linewidth]{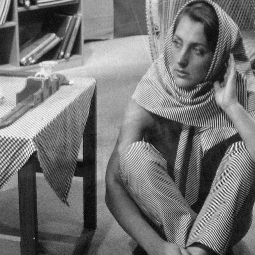}\vspace{4pt}
\includegraphics[width=1\linewidth]{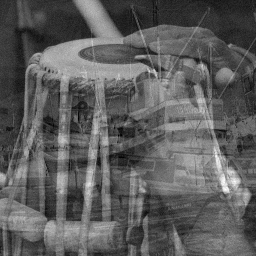}
\end{minipage}}
\subfigure[]{
\begin{minipage}[b]{0.17\linewidth}
\includegraphics[width=1\linewidth]{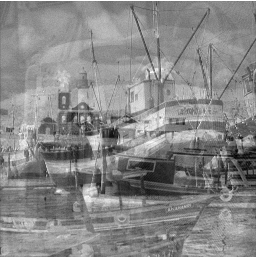}\vspace{4pt}
\includegraphics[width=1\linewidth]{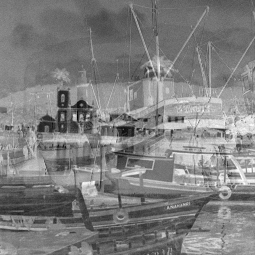}\vspace{4pt}
\includegraphics[width=1\linewidth]{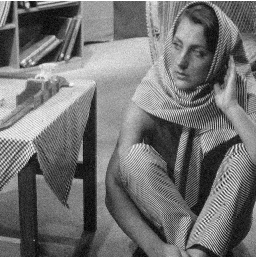}\vspace{4pt}
\includegraphics[width=1\linewidth]{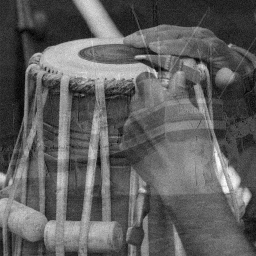}
\end{minipage}}
\subfigure[]{
\begin{minipage}[b]{0.17\linewidth}
\includegraphics[width=1\linewidth]{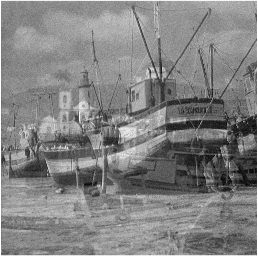}\vspace{4pt}
\includegraphics[width=1\linewidth]{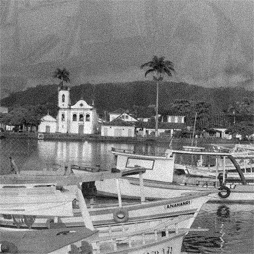}\vspace{4pt}
\includegraphics[width=1\linewidth]{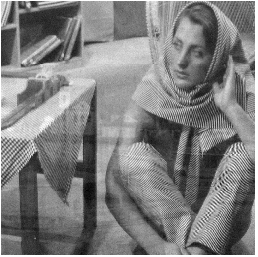}\vspace{4pt}
\includegraphics[width=1\linewidth]{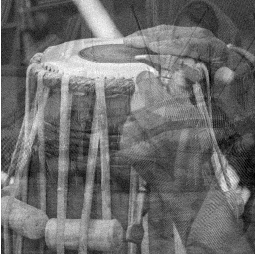}
\end{minipage}}
\subfigure[]{
\begin{minipage}[b]{0.17\linewidth}
\includegraphics[width=1\linewidth]{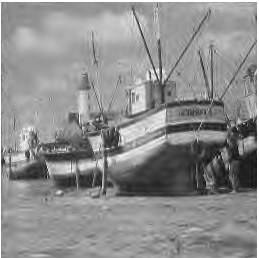}\vspace{4pt}
\includegraphics[width=1\linewidth]{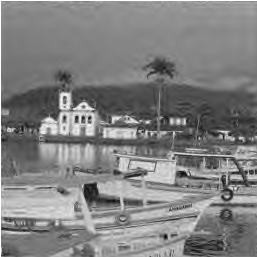}\vspace{4pt}
\includegraphics[width=1\linewidth]{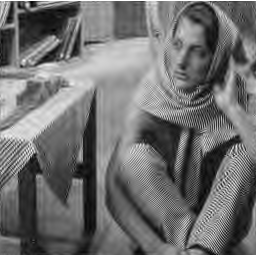}\vspace{4pt}
\includegraphics[width=1\linewidth]{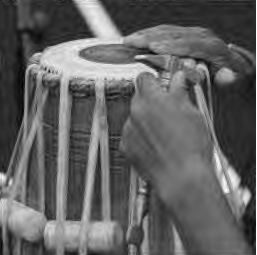}
\end{minipage}}
\caption{Experiment2: image separation (20dB noise); \textbf{(a)}:Original image sources; \textbf{(b)}:EFICA outputs; \textbf{(c)}:JADE outputs; \textbf{(d)}:MMCA outputs; \textbf{(2)}:GMCA outputs.}
\label{BSS_ex2_11}
\end{figure}

GMCA has been proved elsewhere \cite{BobinJ_2007SaMD} that it achieves better separation quality when the estimated components are 'very' sparse in the given dictionary. Therefore GMCA is supposed to perform well on separating images even without distinct discrepancies. Separated images are displayed in Figure (\ref{BSS_ex2_11}). It shows the recovered image by various BSS methods contaminated by 20dB noise. All methods can distinguish the barbara image (row 3) from the mixtures. The ICA methods, nonetheless fail to separate the scenery photos (row 1 and 2). But in (e) using GMCA still gives acceptable result. MMCA using curvelet + DCT in (d) gives better result than ICA methods, but not so good as GMCA.\\

% ---------- begin figure-------------

% ----------end figure------------- %

Figure (\ref{BSS_EV21}) portrays the evolution of average correlation coefficient over 4 estimated sources as a function of the noise variance. At a first glance, GMCA significantly outperforms the ICA methods in terms of robustness and separation quality. Moreover, JADE performs the worst in among all ICA based algorithms. Figure (\ref{BSS_EV21}) depicts the behavior of the mixing matrix criterion as the noise decreases. The mixing matrix criterion also clearly revels the strength of GMCA method.\\

% ----------begin figure------------- %
\begin{figure}[!htbp]
\centering
\begin{minipage}[b]{0.49\textwidth}
\includegraphics[width=\textwidth]{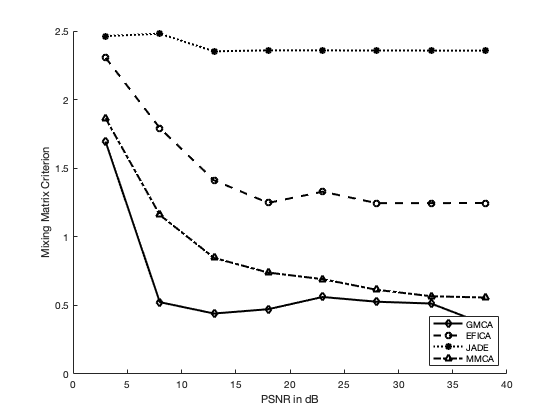}
\caption{Evolution of the mixing matrix criterion (after indeterminacy corrected) as the noise variance varies.}
\label{BSS_EV21}
\end{minipage}
\begin{minipage}[b]{0.49\textwidth}
\includegraphics[width=\textwidth]{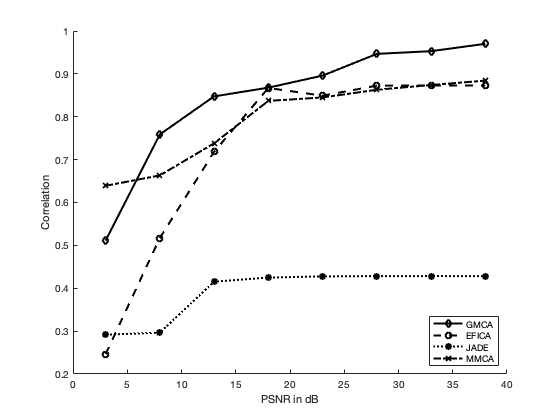}
\caption{Evolution of the correlation coefficient between original and estimated sources as the noise variance varies.}

\label{BSS_EV22}
\end{minipage}
\end{figure}
% ----------end figure------------- %

To summarise the findings in this experiment, GMCA does take good advantage of overcompleteness and morphological diversity. We can aslo claim that sparsity brings better results. Taking the advantages further, in next experiment, we explore how locally learned dictionary helps the blind source separation.\\

\subsection{Blind image separation using adaptive dictionary learning}

In this experiment, simulations are provided to demonstrate the performance of the proposed BK-SVD algorithms, as compared with the baseline algorithms, K-SVD. Same as last experiment settings, a severe case of 4 image sources with very different morphologies was chosen to exam the performance of the methods. 400 iterations were selected as the stopping criterion, additive Gaussian noise is added from 3dB to 40dB. The wavelet based GMCA is selected as the baseline algorithm.\\

\begin{figure}[!htbp]
\centering
\subfigure[]{
\begin{minipage}[b]{0.99\linewidth}
\centering
\includegraphics[width=0.15\linewidth]{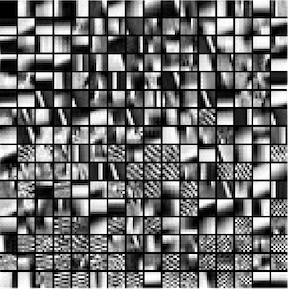}\vspace{4pt}
\includegraphics[width=0.15\linewidth]{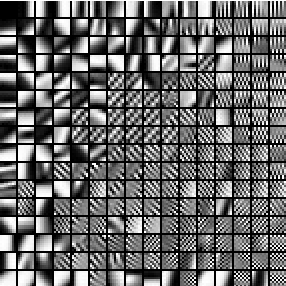}\vspace{4pt}
\includegraphics[width=0.15\linewidth]{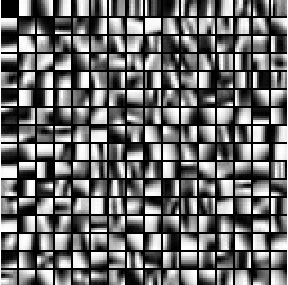}\vspace{4pt}
\includegraphics[width=0.15\linewidth]{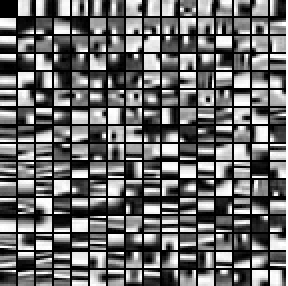}
\end{minipage}}
\subfigure[]{
\begin{minipage}[b]{0.99\linewidth}
\centering
\includegraphics[width=0.15\linewidth]{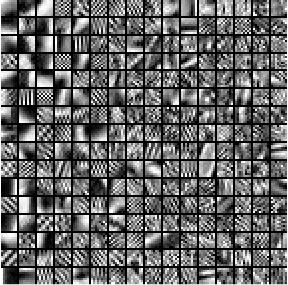}\vspace{4pt}
\includegraphics[width=0.15\linewidth]{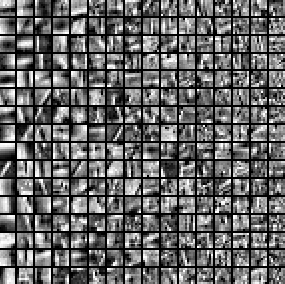}\vspace{4pt}
\includegraphics[width=0.15\linewidth]{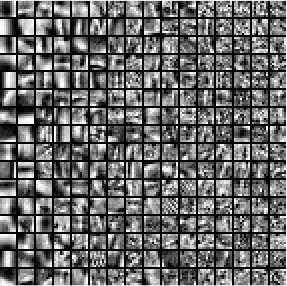}\vspace{4pt}
\includegraphics[width=0.15\linewidth]{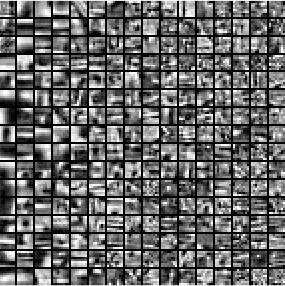}
\end{minipage}}
% \vspace{-0.5cm} 
\caption{(a): Trained dictionary using K-SVD for sources 1,2,3,4 respectively; (b): Trained dictionary using SAC+BK-SVD for the 4 sources.}
\label{dictionary_learned}
\end{figure}

In order to obtain enough training samples for dictionary learning, multiple overlapped segments (patches) of the sources are taken \cite{VAbolghasemi2012}. Choosing the optimal patch size is a subtle problem. Generally, very large patches should be avoided as they lead to massive dictionaries and also provide few training samples for the dictionary learning stage. We choose $8\times 8$ patches for this experiment. Furthermore the patches are $50\%$ overlapped as suggested in \cite{VAbolghasemi2012}. Maximal block size is set to be $s= 3$ and the block sparsity is $k=2$ for BK-SVD+SAC. A standard DCT dictionary is chosen during initialisation. All dictionaries obtained have size of $64\times 256$. Consequently, each dictionary in (\ref{dictionary_learned}) have $16 \times 16 = 256$ blocks whereas each block is obtained from a $8\times 8$ patches. Figure (\ref{dictionary_learned}) also illustrates that both K-SVD and the proposed BK-SVD have good adaption to the corresponding sources, and looks significantly different from the standard DCT dictionary.\\

\begin{figure*}[!htbp]
\centering
\subfigure[]{
\begin{minipage}[b]{0.17\linewidth}
\includegraphics[width=1\linewidth]{images/boat_ori.png}\vspace{4pt}
\includegraphics[width=1\linewidth]{images/paraty_ori.png}\vspace{4pt}
\includegraphics[width=1\linewidth]{images/barbara_ori.png}\vspace{4pt}
\includegraphics[width=1\linewidth]{images/pakhawaj_ori.png}
\end{minipage}}
\subfigure[]{
\begin{minipage}[b]{0.17\linewidth}
\includegraphics[width=1\linewidth]{images/gmca_out1.png}\vspace{4pt}
\includegraphics[width=1\linewidth]{images/gmca_out2.png}\vspace{4pt}
\includegraphics[width=1\linewidth]{images/gmca_out3.png}\vspace{4pt}
\includegraphics[width=1\linewidth]{images/gmca_out4.png}
\end{minipage}}
\subfigure[]{
\begin{minipage}[b]{0.17\linewidth}
\includegraphics[width=1\linewidth]{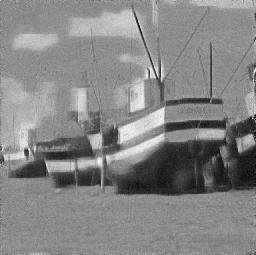}\vspace{4pt}
\includegraphics[width=1\linewidth]{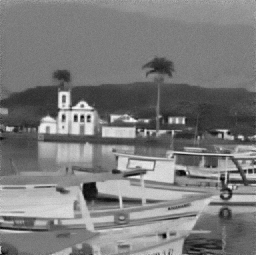}\vspace{4pt}
\includegraphics[width=1\linewidth]{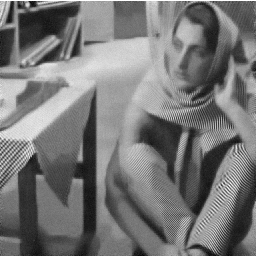}\vspace{4pt}
\includegraphics[width=1\linewidth]{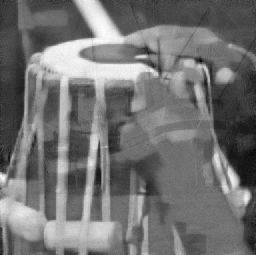}
\end{minipage}}
\subfigure[]{
\begin{minipage}[b]{0.17\linewidth}
\includegraphics[width=1\linewidth]{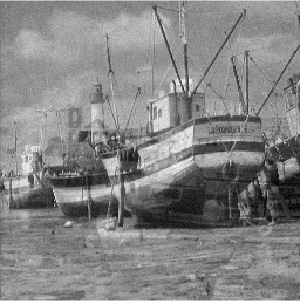}\vspace{4pt}
\includegraphics[width=1\linewidth]{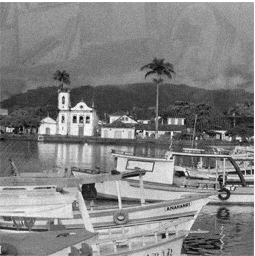}\vspace{4pt}
\includegraphics[width=1\linewidth]{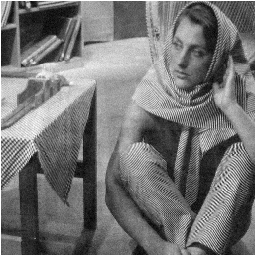}\vspace{4pt}
\includegraphics[width=1\linewidth]{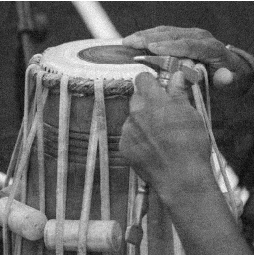}
\end{minipage}}
\caption{Experiment 3 (20dB noise): Using adaptive dictionary learning; \textbf{(a)}: original sources; \textbf{(b)}:GMCA outputs; \textbf{(c)}:K-SVD+MMCA output; \textbf{(d)}:BK-SVD+SAC+MMCA outputs;}
\label{apt_compare}
\end{figure*}

Figure (\ref{apt_compare}) visually compares the separation results
using several algorithm under 20dB Gaussian noise ($\sigma = 15$). Visually we can see that both two adaptive dictionary learning methods outperform the GMCA method in terms of recovered image quality. The learned dictionary helps to restore more details of the original image. Moreover, compared to the proposed methods, separated image in K-SVD+MMCA is a bit blurry. We think 400 iterations maybe too large for this problem and the 
algorithm overfits. Afterall, the proposed method will show superiority to GMCA and K-SVD+MMCA in next paragraph.\\

% ---------begin figure ------------ %
% ---------end figure ------------ %

\begin{figure}[!htbp]
\centering
\begin{minipage}[]{0.49\textwidth}
\includegraphics[width=\textwidth]{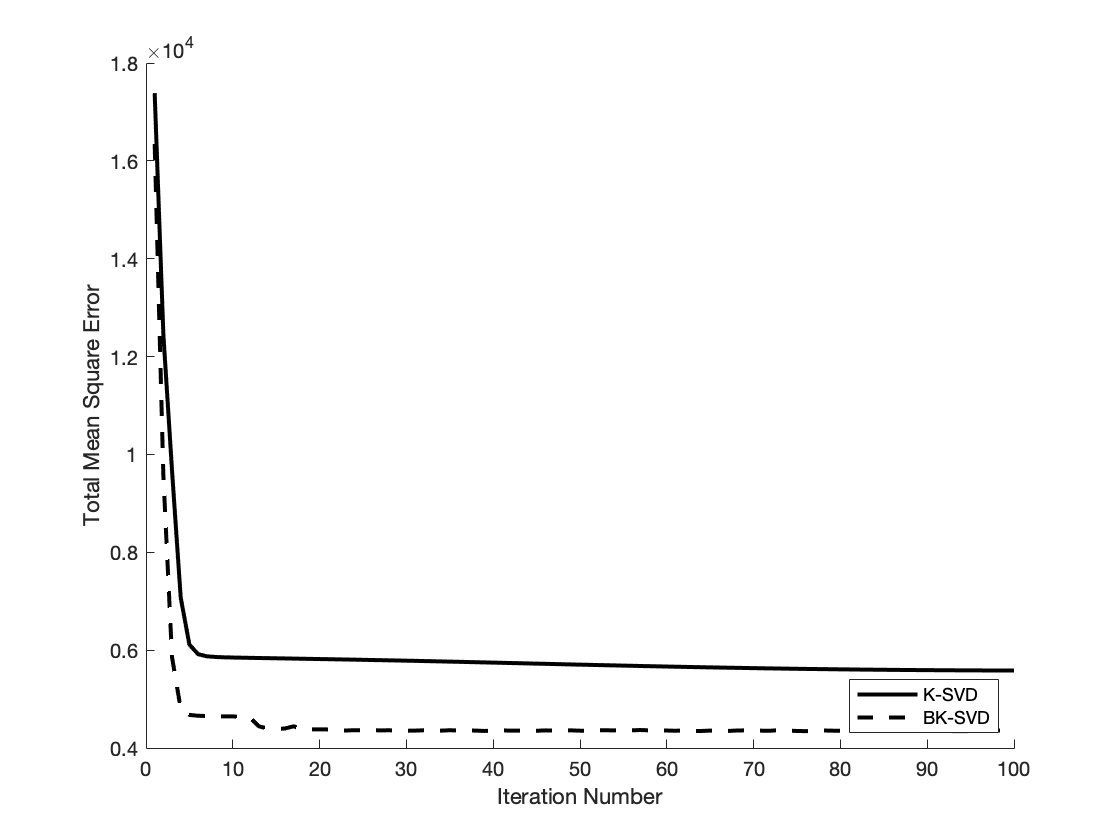}
\caption{Total MSE versus number
of iterations, Note that the elements of
image sources have amplitude in the range $[0, 255]$.}
\label{BSS_EV21_1}
\end{minipage}
\begin{minipage}[]{0.49\textwidth}
\includegraphics[width=\textwidth]{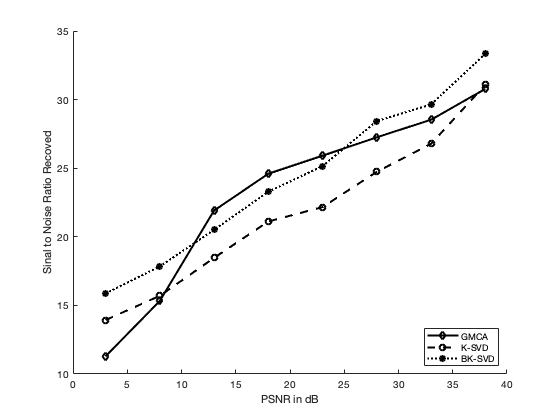}
\caption{Average PSNR of the estimated sources as the Gaussian disturbance changes.}
\label{BSS_EV22_2}
\end{minipage}
\end{figure}

We have shown via visual results that the block-structure dictionary learning algorithm provides convincing contribution to the blind source separation. Moreover, both the correlation and representation error proves our dictionary design method gives better performance. In Figure (\ref{BSS_EV21}), for PSNR smaller than 13dB, the proposed block-sparsifying BSS method yields similar correlation coefficient as the K-SVD method and GMCA. This is because when the SNR is low, the algorithm may no longer successfully build a appropriate block structure. For low noise settings (PSNR is high), the proposed method clearly outperforms the other two methods. All three methods performs well in solving the mixing matrix in low noise settings, where BK-SVD behaves slightly better. Furthermore, Figure (\ref{BSS_EV21_1}) plots the total reconstruction error computed as $||\mathbf{Y} - \mathbf{AX}||_2$. We can see that the proposed SAC+BK-SVD leads to smaller errors than the K-SVD method.\\

% ----------begin figure------------- %
\begin{figure}[!htbp]
\centering
\begin{minipage}[b]{0.49\textwidth}
\includegraphics[width=\textwidth]{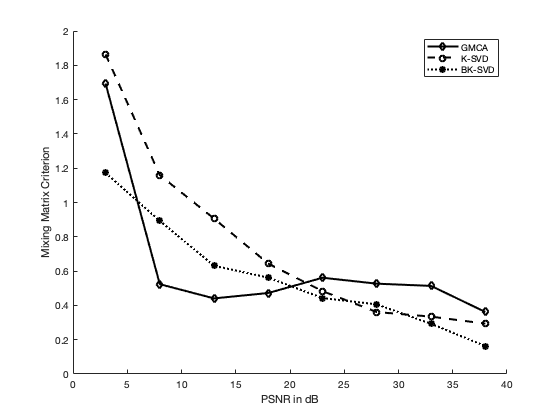}
\caption{Evolution of the correlation coefficient between original and estimated sources as the noise variance varies.}
\label{BSS_EV21}
\end{minipage}
\begin{minipage}[b]{0.49\textwidth}
\includegraphics[width=\textwidth]{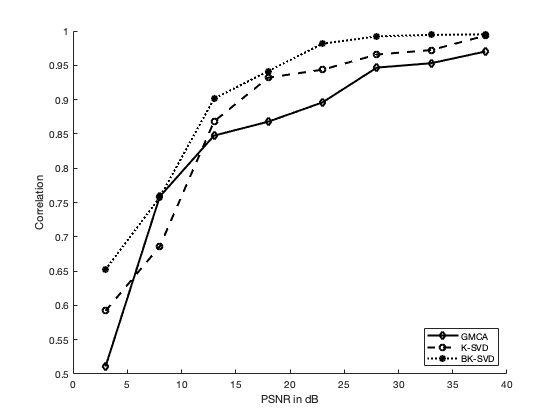}
\caption{Evolution of the mixing matrix criterion (after indeterminacy corrected) as the noise variance varies.}
\label{BSS_EV22}
\end{minipage}
\end{figure}
% ----------end figure------------- %

% ----------begin figure------------- %
% ----------end figure------------- %
The running time for K-SVD BSS algorithm is 238 seconds. However the running time for BK-SVD BSS algorithm is 389 seconds. For Fast GMCA it is only 4 seconds. This convinces our deduction before, as the sparse agglomerative clustering algorithm dominates the complexity. Even though the BK-SVD algorithm is supposed to be faster than K-SVD, but the clustering stage slows down the overall process. It is also seen that, due to learning the dictionary, both algorithms are computationally demanding for large scale applications (i.e. image processing) and are much slower than the GMCA algorithm. This implies that further effort is required to speed up the dictionary learning part in BSS. \\
% --------------begin table ----------- %
\begin{table}[!htbp]
\centering
    \begin{tabular}{|c|c|c|c|}
    \hline
    MMCA & Fast GMCA & K-SVD & BK-SVD+SAC \\\hline
    $221s$ & $4s$ & $238s$ & $389s$ \\\hline
    \end{tabular}
    \caption{Running time for GMCA, KSVD and SAC+BK-SVD respectively, up to 100 iterations}
\end{table}
% --------------end table ----------- %

In this section we examed the proposed BSS framework for the design of a block-sparsifying dictionary given a set of images and a maximal block size. Reuslts shown that it outperforms the prevailing K-SVD BSS algorithm by separation quality and reconstruction accuracy.

\subsection{Choosing the best maximal block size and block sparsity level}
From our experience in last section, we found that finding the optimal selection of best maximal block size $s$ and block sparsity level (number of atoms in a block) $k$ in SAC + B-KSVD is an unclear problem which needs more investigation. To avoid the long running time in BSS framework, we reduce the problem to a simple image reconstruction problem, where an image is given and the algorithm is intended to learn the dictionary and representation coefficients. Reconstruction quality is assessed by the mean square error (MSE). Similar as last experiment, we extract $8 \times 8$ patches from a $256 \times 256$ barbara image and set the maximum number of iterations to be 100. A $64 \times 96$ dictionary is learned and the sparse representation has dimension of $96\times1024$. Hence we can recover and resize the $64\times1024$ image. The MSE result of the proposed methods is again, compared with K-SVD.\\

\begin{figure}[!htbp]
\centering
\begin{minipage}[]{0.49\textwidth}
\includegraphics[width=\textwidth]{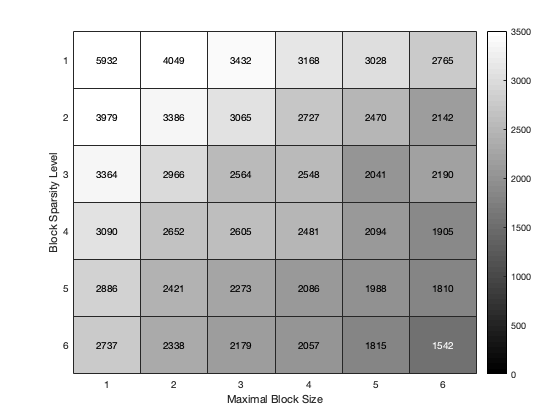}
\caption{K-SVD reconstruction error heatmap against block size and sparsity level.}
\label{HEAT1}
\end{minipage}
\begin{minipage}[]{0.49\textwidth}
\includegraphics[width=\textwidth]{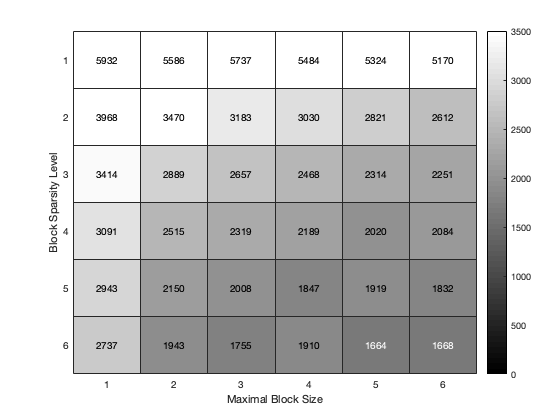}
\caption{BK-SVD reconstruction error heatmap against block size and sparsity level.}
\label{HEAT2}
\end{minipage}
\end{figure}

We vary $s$ and $k$ from 1 to 6 and plot the reconstruction error using two methods in heatmaps. It is obvious that there exists a general trend that the reconstruction error decreases as the block size gets larger. It is reasonable because the more atoms are collected, the better will be the reconstruction quality. But the decreasing error trend is not monotonic for SAC+BK-SVD in Figure (\ref{HEAT2}). For instance, the selection of $s=4$ and $k=5$ gives better outcome than $s=5$ and $k=5$. This means blindly increasing the parameters (more atoms numbers) is not an wise choice in SAC+BK-SVD. Furthermore, for $s\geq 6$ the superiority of the proposed methods no longer holds. To conclude, selecting block size and sparsity level needs discussion on a case by case basis, and is sometimes compromised with the runing time.
% ----------begin figure------------- %

\section{Conclusion and futurework}
In this report, different approaches for blind source separation are investigates and discussed. Then a sparse clustering based block dictionary learning algorithm is applied to the BSS problem. In every iteration of the BSS process, the proposed algorithm repeat two stages, a block structure clustering step (SAC) and a dictionary update setp (BK-SVD). When the maximal block size in SAC is reduced to 1, the proposed algorithm reverts to normal K-SVD. In contrasting to the normal K-SVD dictionary learning BSS algorithm, the proposed one is noted to give a sparser representation of the target image and exhibit a better estimation of the mixing matrix and sources.\\

However the proposed methods has certain limitations. The computation cost of our proposed method is not satisfactory. This is due to the cubed complexity of the SAC algorithm and blockwise updating manner of BK-SVD algorithm. In the future, we may consider the SimCo method \cite{6340354} in computing the dictionary atoms. SimCo allows updating all codewords and all sparse coefficients simultaneously and is expected to significantly speed up the atom updating process. To further improve the proposed adaptive BSS methods, one could try and make the dictionary learning step less susceptible to local minimum traps. In addition, training one dictionary to sparsely represent all the sources is an alternative to calculating multiple distinct dictionaries, as long as the dictionary redundancy is large enough. An obvious advantage of using one dictionary is that the computational cost does not increase when the number of sources increases. Another refinement could be replacing blocks in the dictionary that contributes little to signal representations with the least significant signal elements. This is expected to further improve the reconstruction ability of out dictionary learning algorithm. Besides, extending the proposed block-sparse BSS framework to underdetermined blind separation cases is also a valuable research direction.

\newpage
\bibliography{citations}
\bibliographystyle{apalike}
\end{document}